\documentclass[10pt,journal,compsoc]{IEEEtran}
\ifCLASSOPTIONcompsoc
  \usepackage[nocompress]{cite}
\else
  \usepackage{cite}
\fi
\ifCLASSINFOpdf
\else
\fi

\usepackage{graphicx}
\usepackage{subcaption}
\usepackage{multirow}
\usepackage{xcolor}
\usepackage{footmisc}

\newcommand{\subparagraph}{}
\usepackage[explicit]{titlesec}

\titleformat{\paragraph}[runin]
  {\normalfont\normalsize\bfseries}{}{15pt}{\theparagraph#1.}
  \titleformat{name=\paragraph,numberless}[runin]
  {\normalfont\normalsize\bfseries}{}{15pt}{#1.}
\titlespacing*{\paragraph}{0pt}{0ex plus 0ex minus 0ex}{15pt}

\usepackage{amsmath}
\usepackage{algorithmic}
\usepackage{array}
\usepackage{multirow}
\usepackage{microtype}
\usepackage{subcaption}
\usepackage{graphicx}
\usepackage{makecell}
\usepackage[colorlinks,urlcolor=blue]{hyperref}

\newcommand{\similarity}{s}

\usepackage{url}
\begin{document}
\title{Large-Scale Historical Watermark Recognition: dataset and a new consistency-based approach}

\author{Xi~Shen,
        Ilaria~Pastrolin,
        Oumayma~Bounou,
        Spyros~Gidaris,
        Marc~Smith, 
        Olivier~Poncet,
        Mathieu~Aubry
\IEEEcompsocitemizethanks{\IEEEcompsocthanksitem Xi Shen,  and Mathieu Aubry are at LIGM (UMR 8049), \'{E}cole des Ponts ParisTech, UPE, Marne-la-Vallée, France.\protect\\
E-mail: \{ xi.shen and mathieu.aubry \}@enpc.fr
\IEEEcompsocthanksitem Ilaria Pastrolin, Oumayma Bounou, Marc Smith and Olivier Poncet are at \'{E}cole Nationale des Chartes in France.\protect\\
E-mail: \{ ilaria.pastrolin, oumayma.bounou, marc.smith, olivier.poncet \}@chartes.psl.eu

\IEEEcompsocthanksitem Spyros Gidaris is with Valeo AI in France.\protect\\
E-mail: spyros.gidaris@valeo.com
}
}


\IEEEtitleabstractindextext{%
\begin{abstract}
Historical watermark recognition is a highly practical, yet unsolved challenge for archivists and historians. With a large number of well-defined classes, cluttered and noisy samples, different types of representations, both subtle differences between classes and high intra-class variation, historical watermarks are also challenging for pattern recognition. In this paper, overcoming the difficulty of data collection, we present a large public dataset with more than 6k new photographs, allowing for the first time to tackle at scale the scenarios of practical interest for scholars: one-shot instance recognition and cross-domain one-shot instance recognition amongst more than 16k fine-grained classes. We demonstrate that this new dataset is large enough to train modern deep learning approaches, and show that standard methods can be improved considerably by using mid-level deep features. More precisely, we design both a matching score and a feature fine-tuning strategy based on filtering local matches using spatial consistency. This consistency-based approach provides important performance boost compared to strong baselines. 
Our model achieves 55\% 
top-1 
accuracy on our very challenging 16,753-class one-shot cross-domain recognition task, each class described by a single drawing from the classic Briquet catalog. In addition to watermark classification, we show our approach provides promising results on fine-grained sketch-based image retrieval. 
\end{abstract}

}

\maketitle
\IEEEdisplaynontitleabstractindextext
\IEEEpeerreviewmaketitle

\IEEEraisesectionheading{\section{Introduction}\label{sec:introduction}}

\begin{figure}[!h!]
\centering
\begin{subfigure}{\linewidth}
\centering
 
 \includegraphics[width=0.32\linewidth]{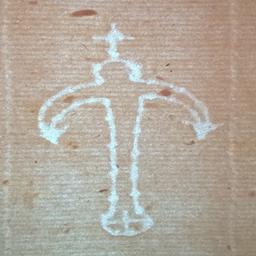}
 \includegraphics[width=0.32\linewidth]{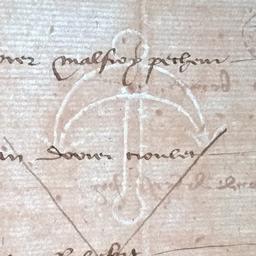}
 \includegraphics[width=0.32\linewidth]{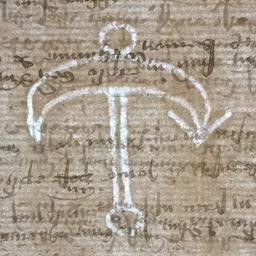}
  \caption{Fine grained recognition (3 different classes)
 \vspace{1mm}}
  \label{fig:teaserb}
 \end{subfigure}
\begin{subfigure}{\linewidth}
\centering
 \includegraphics[width=0.32\linewidth]{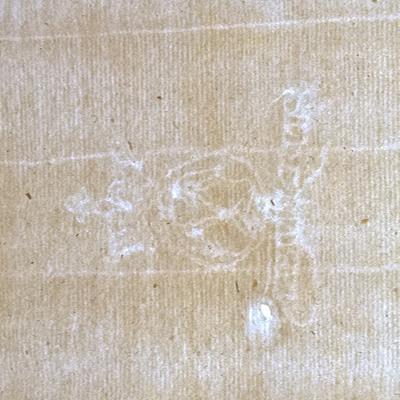}
 \includegraphics[width=0.32\linewidth]{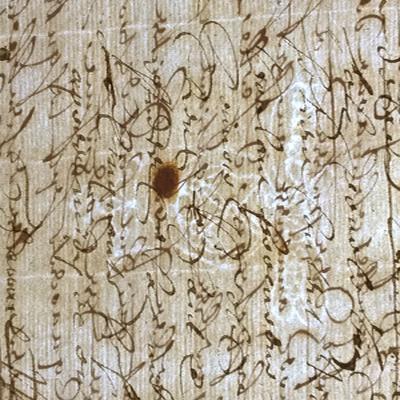}
 \includegraphics[width=0.32\linewidth]{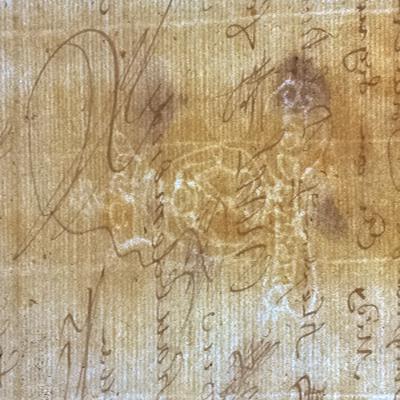}
  \caption{Invariance to appearance and clutter
 \vspace{1mm}}
  \label{fig:teasera}
 \end{subfigure}
\begin{subfigure}{\linewidth}
\centering
 \includegraphics[width=0.32\linewidth]{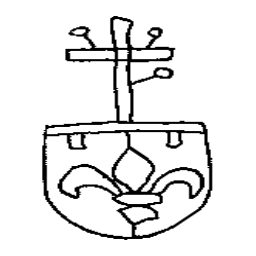}
 \includegraphics[width=0.32\linewidth]{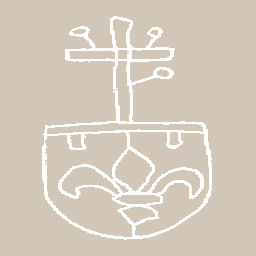}
 \includegraphics[width=0.32\linewidth]{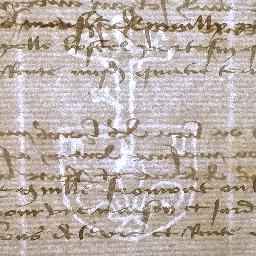}
  \caption{Cross-domain recognition (drawing, synthetic, photograph) }
  \label{fig:teaserc}
 \end{subfigure}
\caption{The key challenges in historical watermark recognition correspond to important problems in computer vision. We collected a dataset to evaluate the performance of state-of-the-art deep learning approaches. Our results show that watermark recognition is a tractable but difficult task, and that leveraging explicitly spatial consistency is key to obtain good results.
}
 \vspace{-2mm}
  \label{fig:teaser}
\end{figure}

\IEEEPARstart{W}{atermark}  instance recognition is a well defined but challenging pattern recognition problem, due to similar categories (Figure \ref{fig:teaserb}), extreme appearance variations (Figure \ref{fig:teasera}) and the necessity to use different types of representations (Figure \ref{fig:teaserc}). 
 In this paper, we demonstrate that leveraging spatial verification of mid-level deep feature matches both at testing and training time leads to clear performance improvement over global state-of-the-art CNN methods. Our approach is an extension and adaptation of the approach we introduced in~\cite{shen2019discovery} for discovering repeated patterns in artwork collections. It is also reminiscent of classical geometric verification of local feature matches \cite{sivic2003video, shen2019discovery}. It can be seen as an argument against the intuition that CNNs rely on high-level arrangements of local patterns, in line with very recent work  showing the performance of CNNs is mainly based on order-less local features aggregation. \cite{brendel2019approximating,geirhos2019imagenet}.

{Recognizing watermarks is also a highly practical problem for archivists and historians since the date and location information associated with each watermark is an important clue to analyse and assess historical documents}. Until now, research on automatic watermark recognition has been held back by the difficulty of creating a large-scale, curated, fine-grained and public dataset. Most proposed techniques have been demonstrated only on a small, non-public and/or very coarse database. We solved this problem by collecting a dataset in public notarial archives, where it was possible: (i)~to access a wide variety of different watermarks from one place and in chronological sequence ; (ii) to have many examples of the same watermark on papers in different states of conservation and written on by different hands. This allowed us to collect more than 6k unique samples for hundreds of classes. Moreover, we searched and photographed systematically the exact original watermarks corresponding to hundreds of line drawings printed in Briquet's classic catalog of approximately 17k watermarks \cite{briquet1907filigranes}, creating a cross-domain database, which allows to directly evaluate a task of high practical interest: cross-domain one-shot fine-grained recognition. Our dataset is the first watermark dataset with such characteristics, allowing for the first time to train and evaluate deep learning algorithms for fine-grained and cross-domain watermark recognition. We demonstrate results that are of high enough quality to be of immediate practical interest, recognizing a watermark from a single photograph amongst more than 16k fine-grained classes described by a single drawing with 55\% top-1 accuracy.\\

The rest of the paper is organized as follows. First we discuss related work in Section \ref{sec:rw}. Second, in Section \ref{sec:datasets}, we present in more detail historical watermarks, the challenges for their recognition and we discuss our dataset characteristics. Third, in Section \ref{sec:method}, we present our method and its motivation. Finally, in Section \ref{sec:results}, we perform an in-depth analysis of our results on watermark recognition, including comparisons with baselines for both one-shot and cross-domain recognition, as well as the results of our method on a different task, sketch-based image retrieval.\\ 

Our main contributions are:
\begin{itemize}
    \item the creation of a large scale fine-grained public dataset for historical watermark recognition, with emphasis on one-shot recognition and cross-domain recognition; 
    \item a new approach to {fine-grained one-shot cross-domain recognition}
    , which relies on explicitly matching mid-level features and leverages spatial consistency to score matches and fine-tune features; 
    \item an analysis of reasons our approach outperforms standard deep baselines, demonstrating that we overcome some of their limitations.

\end{itemize}
Our data and code are available on our project website \footnote{\url{http://imagine.enpc.fr/~shenx/Watermark}}.

\section{Related work}
\label{sec:rw}

\begin{table*}[t!]
  \begin{center}
    \begin{tabular}{|c|c|c|c|c|c|c|}
     \hline
Dataset & Public& Classes & Images per class & Origin & Framing & Focus\\ \hline
\cite{frauenknecht2015wzis}&no& 12 & $\sim 7.5k$ mixed ($\sim 90\%$ drawing)&aggregation&none& large scale coarse categories\\ \hline

\multirow{2}{*}{ours A} &\multirow{2}{*}{yes} & 100 meta& 60  photographs& {notarial archives}& {inside 2:3 box }&  fine-grained categories\\ 
& & 100 test& 1 'clean' + 2 normal photographs&17th century & with context& one-shot classification\\ \hline
 {Briquet}~\cite{briquet1907filigranes,briquetonline}&yes& $\sim 17k$
 &1 drawing& European archives& none& catalog\\ \hline {Briquet-ours}&yes& {16,753}
 &1 drawing& ~\cite{briquet1907filigranes,briquetonline}& inside 2:3 box& recognition database\\ \hline
\multirow{3}{*}{ours B}&\multirow{3}{*}{yes} 
& & associated to a drawing from \cite{briquet1907filigranes}  & \cite{briquet1907filigranes}, & & one-shot\\
&&{140 train}& {1-7 photographs (463 total)} & parisian archives &{inside 2:3 box }&cross domain fine-\\ 
&&{100 test}&{2 photographs}& 14th-16th century& with context &grained recognition  
\\ \hline
    \end{tabular}
    \caption{Comparison of ours and existing datasets for watermark recognition. 
      }
    \label{tab:data}
  \end{center}
\end{table*}

\IEEEPARstart{W}{e} first review methods for watermark representation and recognition, as well as existing datasets. We then review work related to our two main challenges: few-shot and cross-domain recognition and give a brief overview of local feature approaches in recognition, which are the most related to the approach we propose. Finally, we explain the main difference between this work and our previous work ArtMiner~\cite{shen2019discovery}. \\

\paragraph*{Historical watermark imaging and recognition} 
A complete review of techniques developed to reproduce watermarks is outside the scope of this work and can be found for example in \cite{boyle2009watermark}. We will focus on simple approaches: manual tracing and back-lit photography. Manual tracing simply consists in copying the watermark pattern on tracing paper and was used historically to create the main catalogs of watermarks, such as \cite{piccard1977wasserzeichenkartei,briquet1907filigranes}. The most important of these catalogs are aggregated in online databases~\cite{bernstein}, such as~\cite{briquetonline} which includes specifically the drawings from~\cite{briquet1907filigranes} and on which we build. These databases, however, are hard to leverage without considerable expertise, since the watermarks are mainly described by subjective verbal terms.
Back-lit photography is the most common and convenient technique to acquire an actual photograph of a watermark. While a watermark is often barely visible by looking at the light reflected on the paper, placing it in front of a light source and looking at the transmitted light reveals it more or less clearly, depending on the texture and thickness of the paper. This can be done simply by placing the paper in front of the sun or more conveniently using a light-sheet. 

This duality between the drawings available in catalogs and the photographs one would like to identify, is the source of one of the main challenges of watermark recognition: cross-modality. Several studies have thus focused on localizing and extracting the pattern of a watermark from a back-lit photograph, sometimes also exploiting aligned reflected light images \cite{hiary2007system,hiary2008paper,said2016watermark}.

These techniques could potentially be used to help match a photograph to a database of drawings, as proposed in \cite{rauber1997retrieval}. It is, however, difficult to separate watermarks from other lines in the paper. These techniques are thus often complex, with several parameters to tune, and have not yet been demonstrated on a large scale. 
Thus, most work on watermark recognition focuses on drawings. Older studies such as \cite{rauber1997retrieval} use histogram-based descriptors, in a spirit similar to shape context \cite{belongie2001shape}, while more recent work uses machine learning approaches, such as dictionary learning \cite{picard2016non} or neural networks \cite{pondenkandath2018identifying}.

The study most similar to ours is probably \cite{pondenkandath2018identifying}, which used a non-publicly available database~\cite{frauenknecht2015wzis} of approximately 106,000 watermark reproductions (around 90,000 of which are drawings from one of the main watermark catalogs) and trained a convolutional neural network to classify them into 12 coarse categories. While this is proof that CNNs can be used to classify watermarks, it is different from our work in several key aspects. First, the 12 coarse categories do not correspond to a single watermark, and the results of the classification are of little practical interest to identify a specific watermark. On the contrary, each of our classes corresponds to a single watermark (i.e., each drawing from a catalog would correspond to a different class). Second, \cite{pondenkandath2018identifying} does not consider how trained features can generalize to new categories defined by a single example, a key problem for practical applications. Third, we explicitly separate the problem of cross-domain recognition, using a photograph to retrieve an exact watermark using a catalog of drawings. We designed experiments and acquired images specifically to evaluate each task, organized them to be easily used with standard machine-learning frameworks and secured the rights to distribute them publicly.\\

\paragraph*{One-shot recognition}

The easiest deep approach to one-shot recognition~\cite{fei2006one} is to use a nearest-neighbor classifier with ConvNet-based features pre-trained on a different but similar set of categories for which more training data is available. More advanced approaches try to compensate for the lack of training data by employing meta-learning mechanisms that learn how to recognize an object category from a single example.
Similarly to the nearest-neighbor approach described above, this assumes that a larger and similar dataset is available to learn this learn-to-learn meta-task. There is a broad class of meta-learning-based one-shot recognition approaches, including: 
metric-learning-based approaches that, in order to classify a test example to one of the available categories, learn a similarity function between the test example and the available training examples~\cite{vinyals2016matching, snell2017prototypical, koch2015siamese, yang2018learning, wang2018low} or learn how to access a memory module with the training examples~\cite{garcia2017few, mishra2018simple, santoro2016meta, kaiser2017learning, munkhdalai2017meta}; 
approaches that learn how to predict one-shot classifier parameters conditioned on the few available training data~\cite{gidaris2018dynamic, qi2018low, gomez2005evolving, qiao2017few, ha2016hypernetworks}; gradient-descent-based approaches~\cite{ravi2016optimization, finn2017model, andrychowicz2016learning} that learn how to rapidly adapt a model to a given one-shot recognition task with gradient-descent iterations. As baselines, we evaluate two recent meta-learning-based one-shot recognition approaches that have been shown to exhibit state-of-the-art performance, Matching Networks~\cite{vinyals2016matching} and the approach proposed by Gidaris and Komodakis~\cite{gidaris2018dynamic}. \\

\paragraph*{Cross-domain recognition} There are many scenarios in which one would like to classify or search images in one modality using another as reference. Existing datasets include datasets of clean stock photographs and their counterparts in a realistic environment \cite{saenko2010adapting}, datasets of synthetic images and real photographs \cite{peng2017visda} and joint datasets of drawings and photographs \cite{sketchy2016,xu2018sketchmate,yu2016sketch}. Compared to these, the specificity of watermark recognition is that it is a pattern-recognition problem, without any 3D effects, and that the tracings are faithful to the original watermarks. It thus allows to focus on a relatively simple form of the domain-transfer problem, which still proves very challenging.\\
A complete review of cross-domain recognition approaches is outside the scope of this work, a recent survey can be found in \cite{csurka2017domain}. We have selected as baselines three very different types of approaches requiring only a small amount of data. First, we considered an unsupervised approach, in the spirit of  \cite{sun2016return}, aligning the statistics of the source and target domains. Second, we experimented with a supervised approach, directly learning a mapping between source and target features \cite{massa2016deep,rad2018feature}. Finally, we used a randomization approach, learning invariance to the appearance of the watermark by compositing its pattern with random backgrounds, in a spirit similar to \cite{su2015render}.\\

\paragraph*{Local features and recognition} Recent work analyzing the performance of CNNs \cite{brendel2019approximating,geirhos2019imagenet} suggests that they might ignore a large part of the spatial information present in the image, and rather work in a way similar to classical order-less bags-of-features methods \cite{wallraven2003recognition,grauman2006pyramid,zhang2007local}. This might not be suitable for problems such as watermark recognition, where the actual shape of the watermark is key, especially for fine-grained classification. To build on the local features learned by CNNs but consider spatial information, we follow an approach closely related to the classic spatial verification step introduced in Video Google~\cite{sivic2003video} with SIFT features~\cite{lowe2004distinctive}. Rather than using SIFTs, which we found were not adapted to watermarks, we use intermediary deep features, which can be thought of as mid-level image features. Mid-level features~\cite{singh2012unsupervised,doersch2013mid,doersch2014context} have been used in the context of cross-domain matching in \cite{aubry2014painting}. Our feature fine-tuning, which also leverages the spatial structure of images, is related to self-supervised feature learning methods which use spatial information to define an auxiliary task~\cite{doersch2015unsupervised,noroozi2016unsupervised} and to the recent work of~\cite{Rocco18b} which uses neighborhood consensus to learn correspondences from the correlation map, and which we adapted to use as baseline.\\

{\paragraph*{Relation to ArtMiner~\cite{shen2019discovery}} The method we present is an extension of our previous work~\cite{shen2019discovery}, which targets style-invariant pattern mining in artworks. The main differences are that in~\cite{shen2019discovery} the images were not aligned and no annotations were available. Here, we can leverage the fact that the watermarks are coarsely aligned as well as some class-level supervision. Thus, we can restrict positive local matches during feature training to matches in images from the same classes and at similar spatial locations and we can avoid performing RANSAC to compute potential alignments. The watermark recognition problem we present here is also more clearly defined than pattern mining in artworks, with clear domains and ground truth classes.
}

\section{Dataset construction}
\label{sec:datasets}

\begin{figure*}[!h]
	\centering
 	\begin{subfigure}{\linewidth}
 	\begin{center}
		\includegraphics[width=0.075\linewidth]{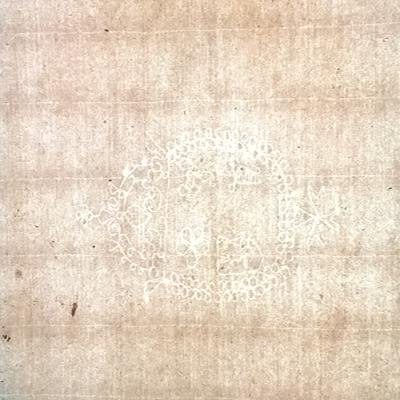}
		\includegraphics[width=0.075\linewidth]{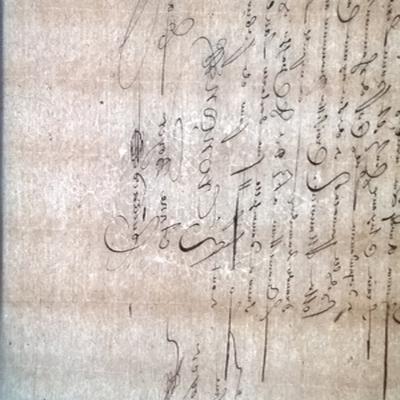}
		\includegraphics[width=0.075\linewidth]{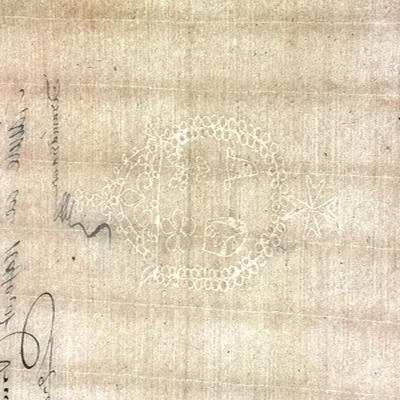}
		\includegraphics[width=0.075\linewidth]{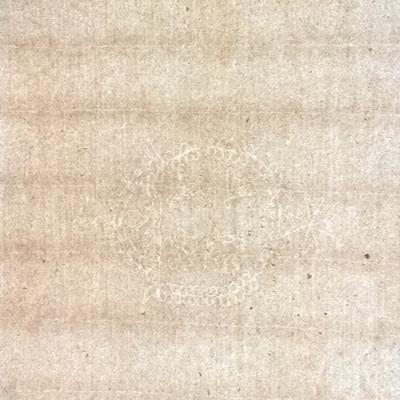}
		\includegraphics[width=0.075\linewidth]{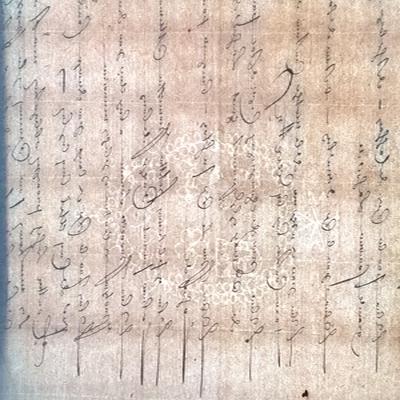}
		\includegraphics[width=0.075\linewidth]{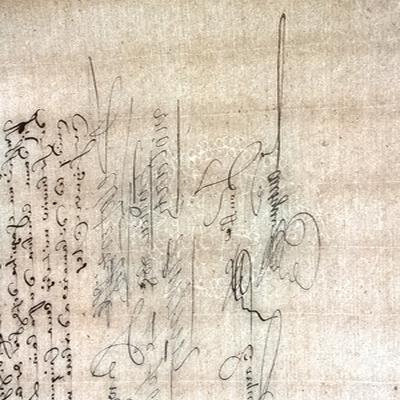}
		\includegraphics[width=0.075\linewidth]{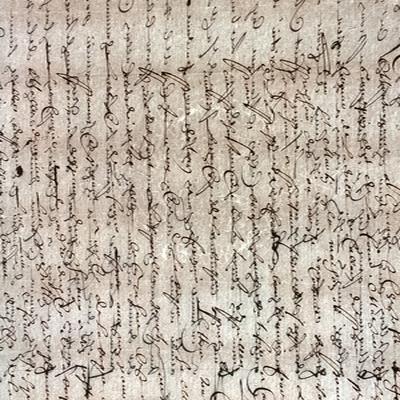}
		\includegraphics[width=0.075\linewidth]{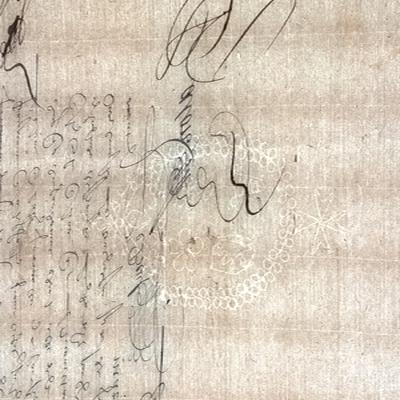}
		\includegraphics[width=0.075\linewidth]{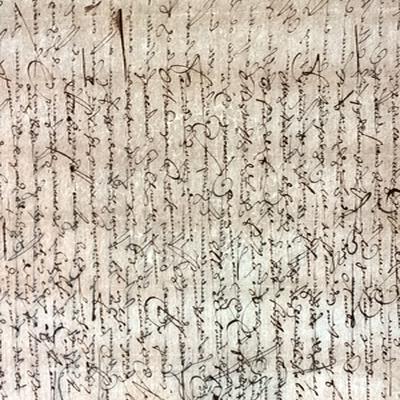}
		\includegraphics[width=0.075\linewidth]{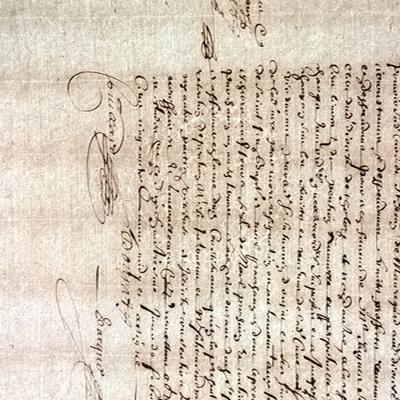}
		\includegraphics[width=0.075\linewidth]{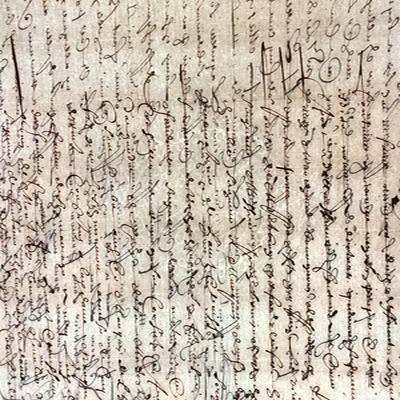}
		\vspace{1mm}
		\includegraphics[width=0.075\linewidth]{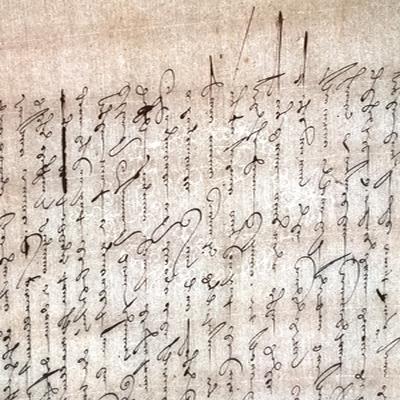}
		\includegraphics[width=0.075\linewidth]{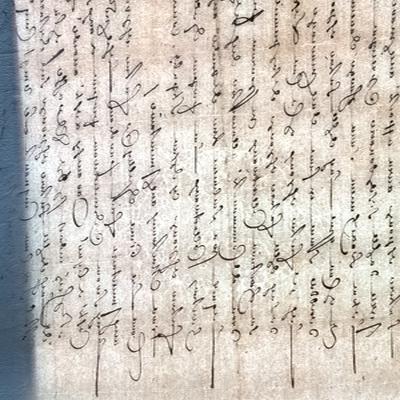}
		\includegraphics[width=0.075\linewidth]{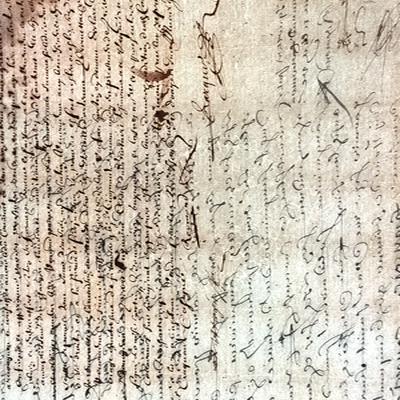}
		\includegraphics[width=0.075\linewidth]{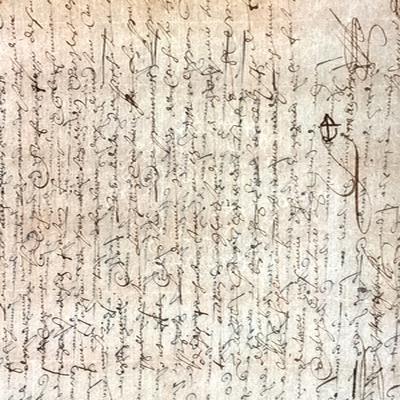}
		\includegraphics[width=0.075\linewidth]{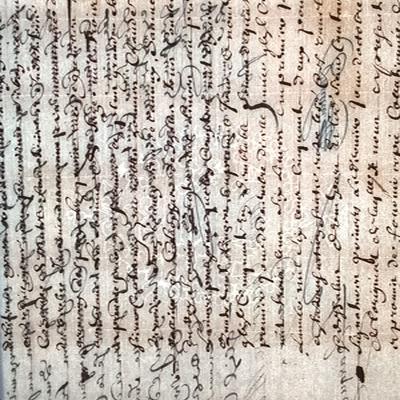}
		\includegraphics[width=0.075\linewidth]{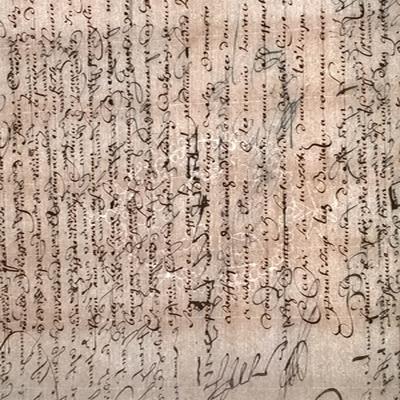}
		\includegraphics[width=0.075\linewidth]{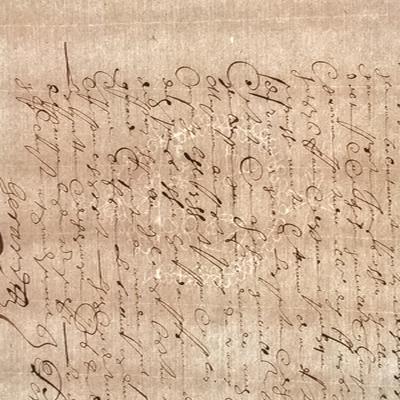}
		\includegraphics[width=0.075\linewidth]{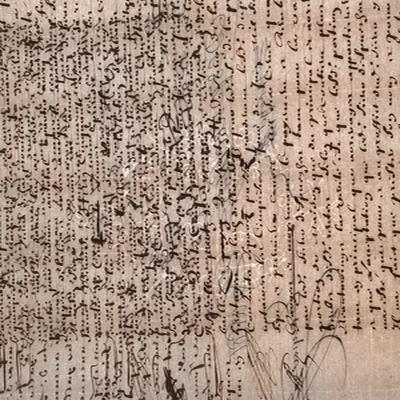}
		\includegraphics[width=0.075\linewidth]{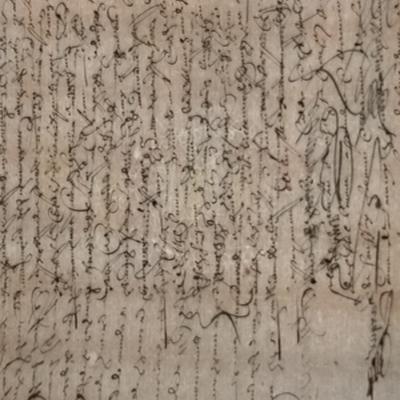}
		\includegraphics[width=0.075\linewidth]{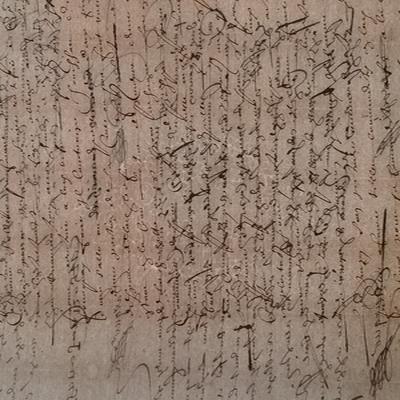}
		\includegraphics[width=0.075\linewidth]{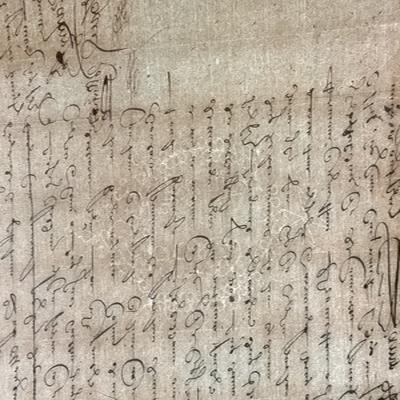}
		\includegraphics[width=0.075\linewidth]{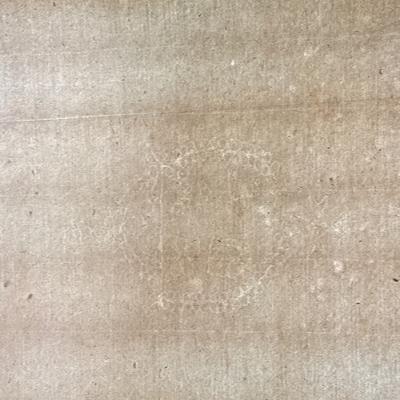}
		\vspace{1mm}
		\includegraphics[width=0.075\linewidth]{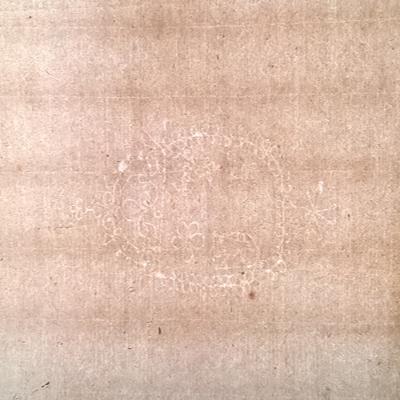}
		\includegraphics[width=0.075\linewidth]{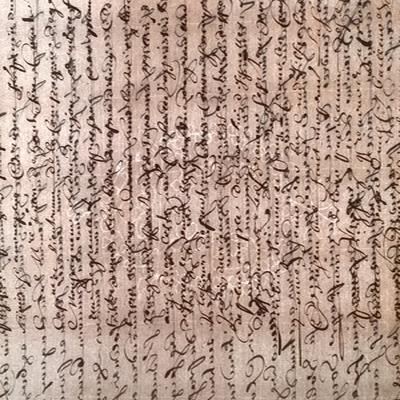}
		\includegraphics[width=0.075\linewidth]{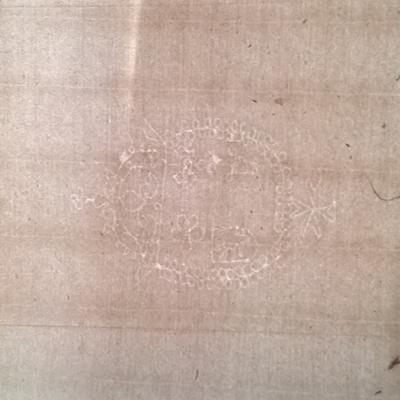}
		\includegraphics[width=0.075\linewidth]{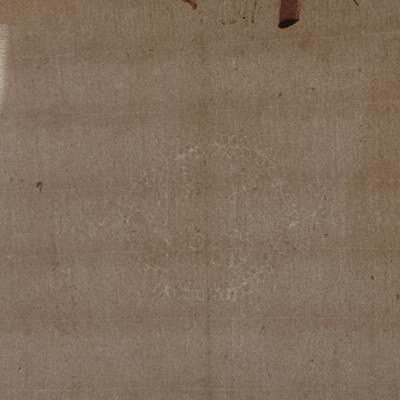}
		\includegraphics[width=0.075\linewidth]{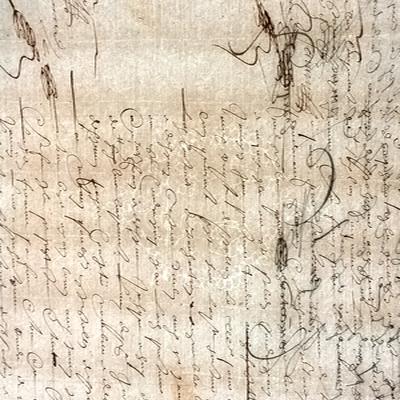}
		\includegraphics[width=0.075\linewidth]{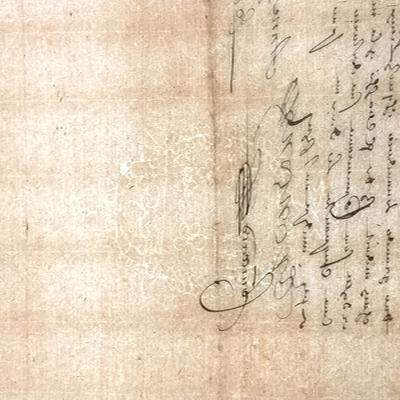}
		\includegraphics[width=0.075\linewidth]{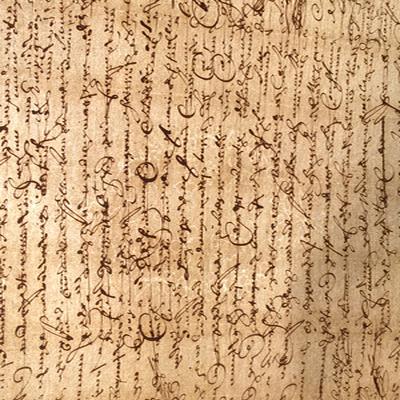}
		\includegraphics[width=0.075\linewidth]{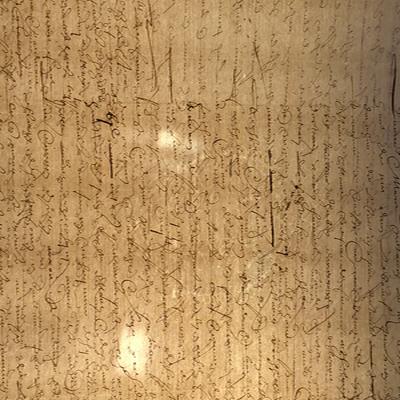}
		\includegraphics[width=0.075\linewidth]{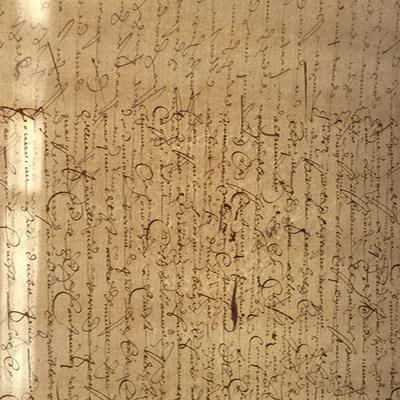}
		\includegraphics[width=0.075\linewidth]{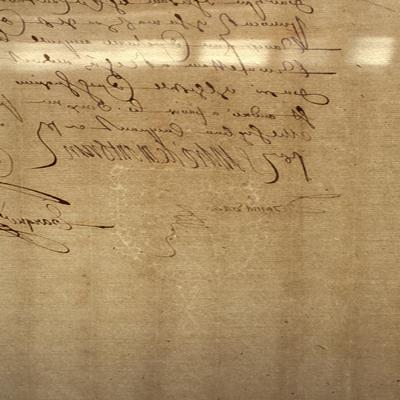}
		\includegraphics[width=0.075\linewidth]{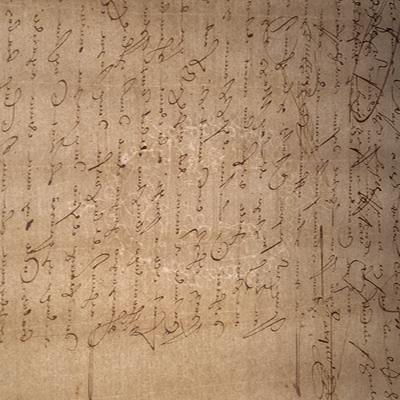}
		\includegraphics[width=0.075\linewidth]{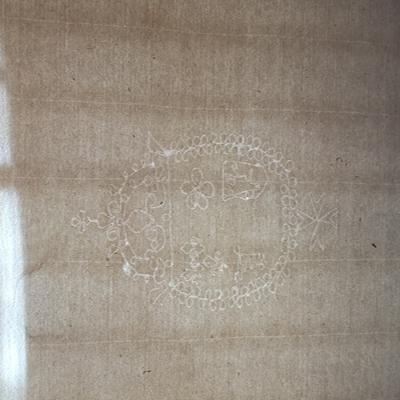}
		\includegraphics[width=0.075\linewidth]{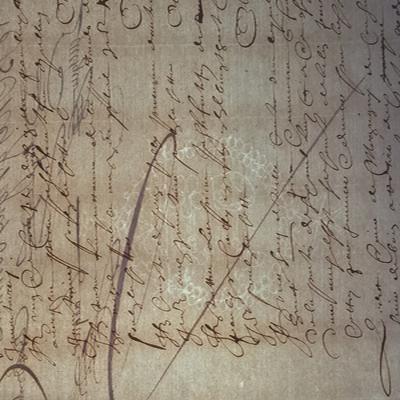}
	\end{center}
	\caption{Different instances of the same watermark from one class of our dataset A. Notice the differences in appearance.}
	\label{fig:Atrain}
 	\end{subfigure}
 	
 	\begin{subfigure}{\linewidth}
 	\begin{center}
		\includegraphics[width=0.075\linewidth]{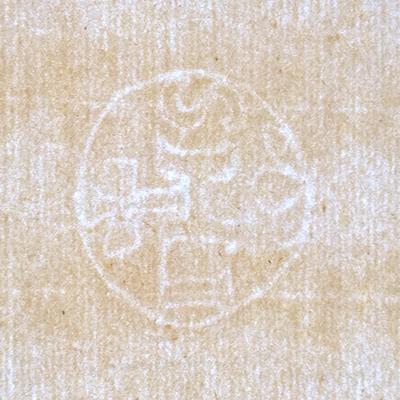}
		\includegraphics[width=0.075\linewidth]{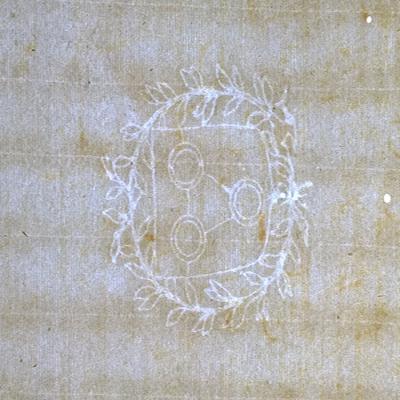}
		\includegraphics[width=0.075\linewidth]{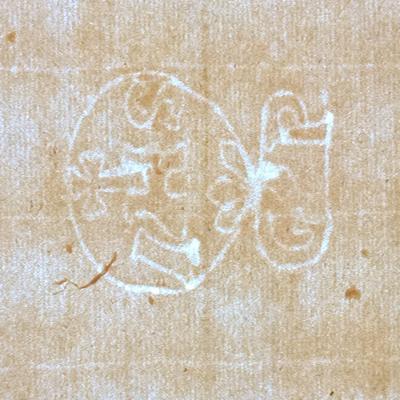}
		\includegraphics[width=0.075\linewidth]{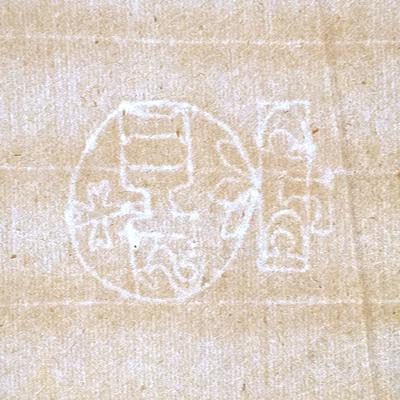}
		\includegraphics[width=0.075\linewidth]{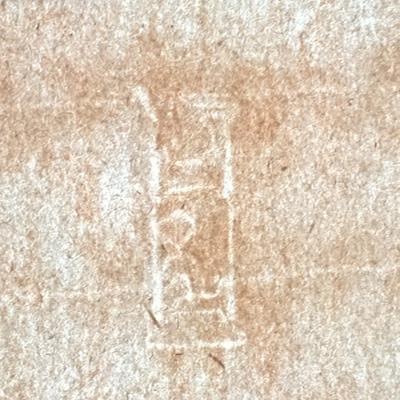}
		\includegraphics[width=0.075\linewidth]{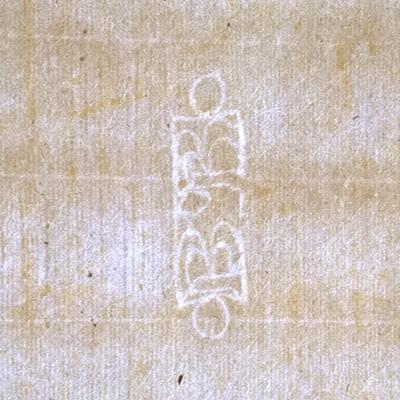}
		\includegraphics[width=0.075\linewidth]{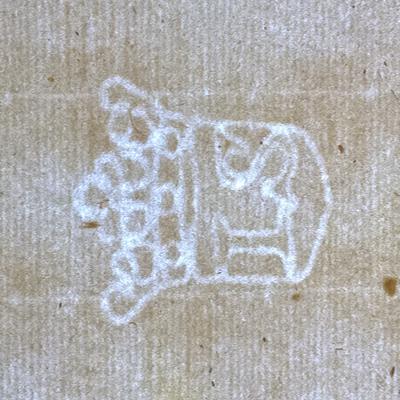}
		\includegraphics[width=0.075\linewidth]{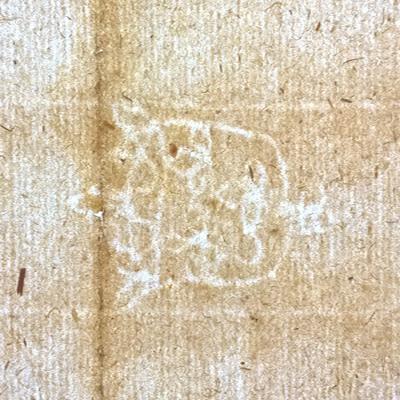}
		\includegraphics[width=0.075\linewidth]{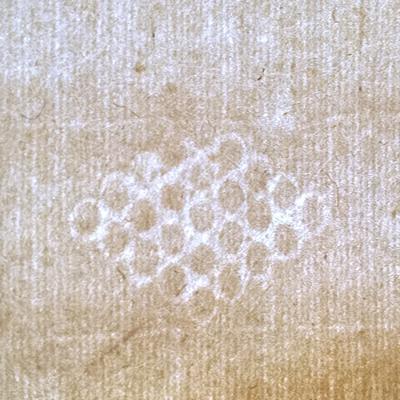}
		\includegraphics[width=0.075\linewidth]{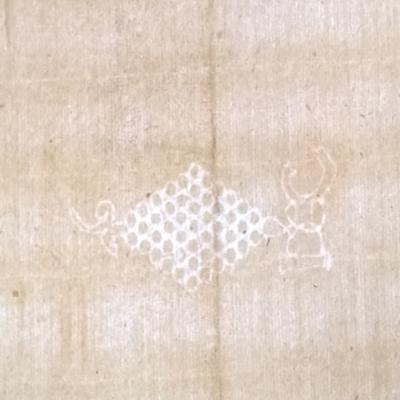}
		\includegraphics[width=0.075\linewidth]{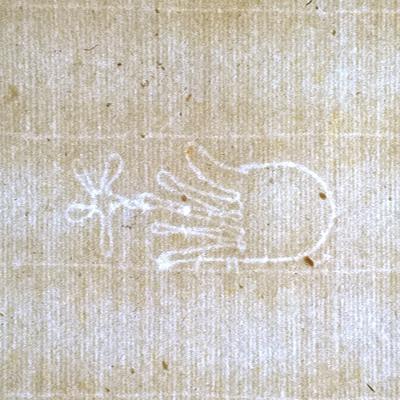}
		\includegraphics[width=0.075\linewidth]{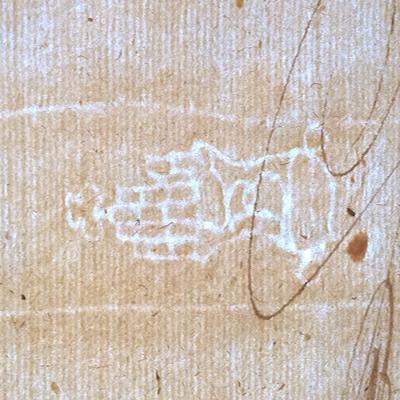}
		\\[1mm]
		\includegraphics[width=0.075\linewidth]{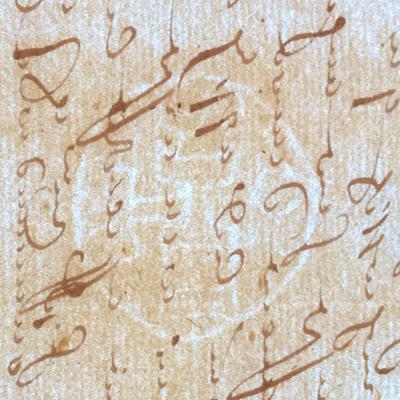}
		\includegraphics[width=0.075\linewidth]{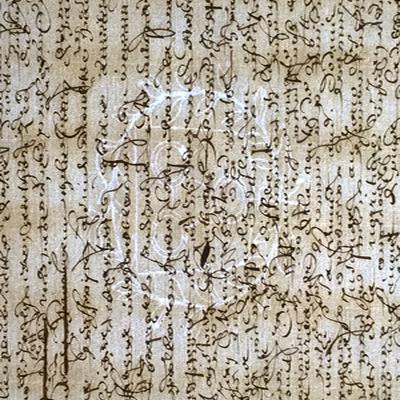}
		\includegraphics[width=0.075\linewidth]{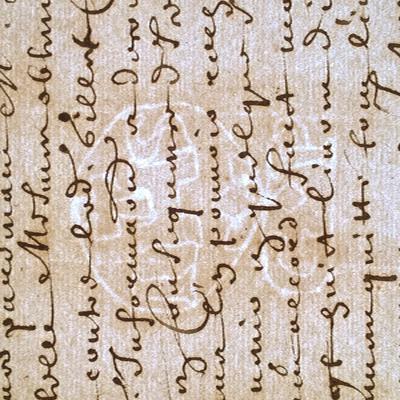}
		\includegraphics[width=0.075\linewidth]{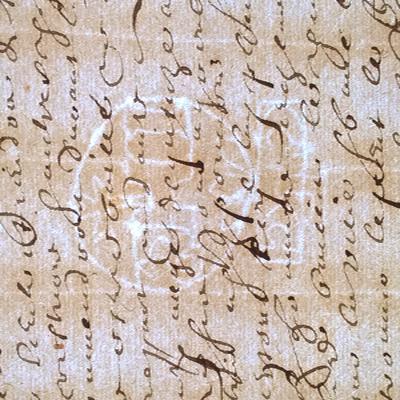}
		\includegraphics[width=0.075\linewidth]{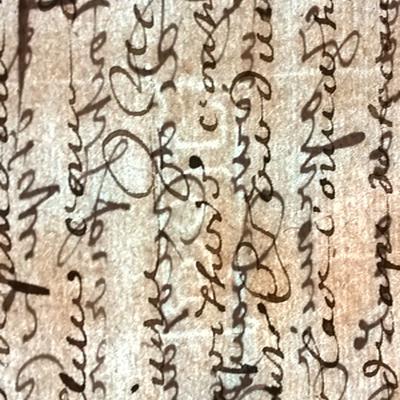}
		\includegraphics[width=0.075\linewidth]{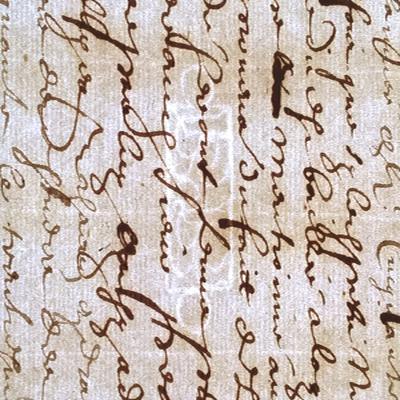}
		\includegraphics[width=0.075\linewidth]{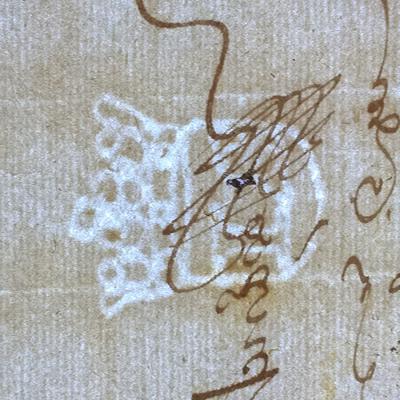}
		\includegraphics[width=0.075\linewidth]{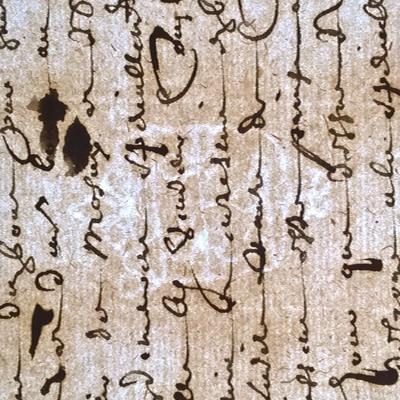}
		\includegraphics[width=0.075\linewidth]{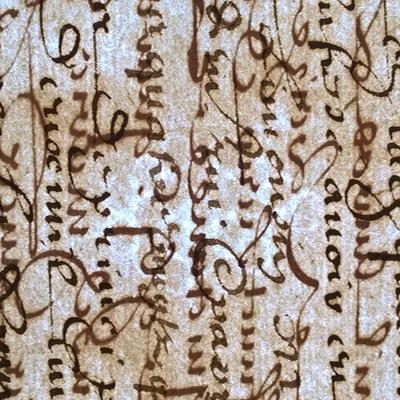}
		\includegraphics[width=0.075\linewidth]{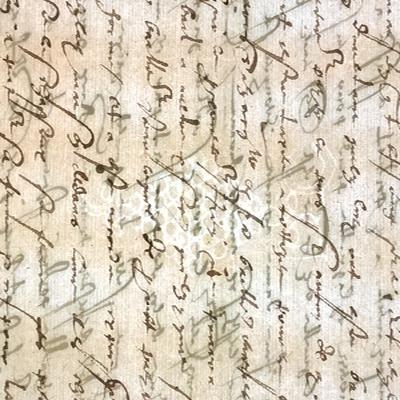}
		\includegraphics[width=0.075\linewidth]{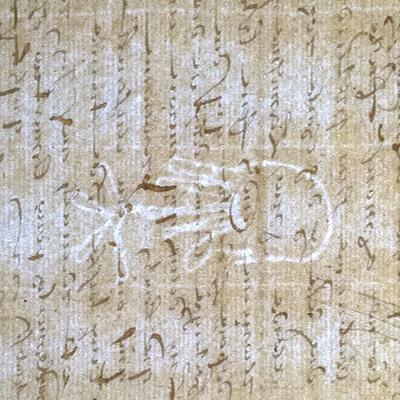}
		\includegraphics[width=0.075\linewidth]{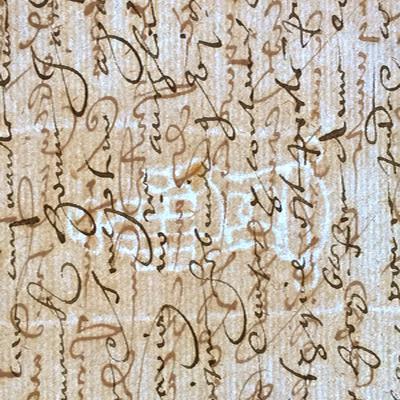}
		\\[1mm]
		\includegraphics[width=0.075\linewidth]{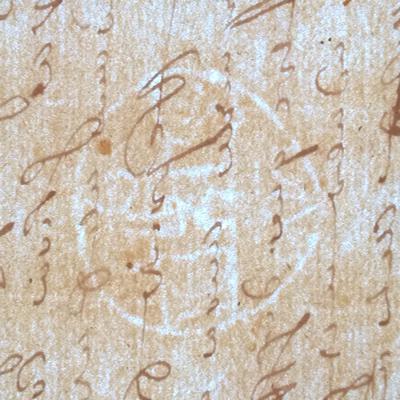}
		\includegraphics[width=0.075\linewidth]{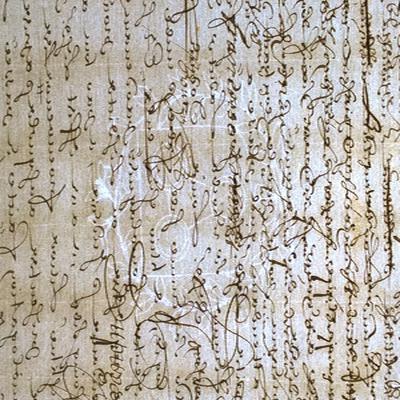}
		\includegraphics[width=0.075\linewidth]{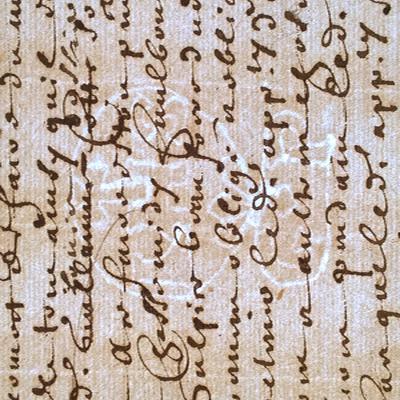}
		\includegraphics[width=0.075\linewidth]{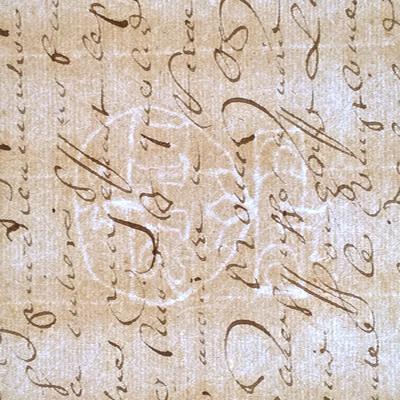}
		\includegraphics[width=0.075\linewidth]{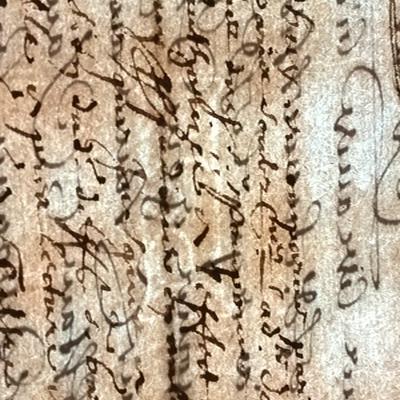}
		\includegraphics[width=0.075\linewidth]{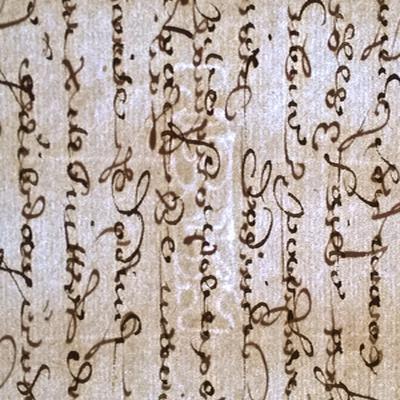}
		\includegraphics[width=0.075\linewidth]{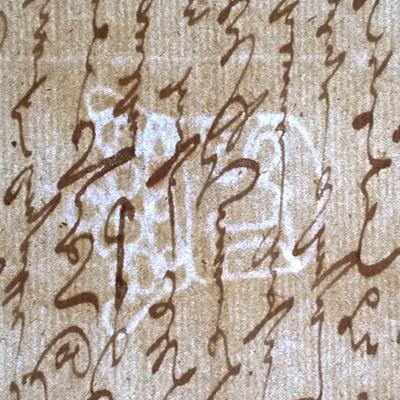}
		\includegraphics[width=0.075\linewidth]{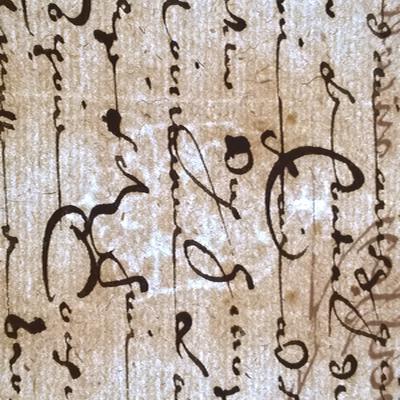}
		\includegraphics[width=0.075\linewidth]{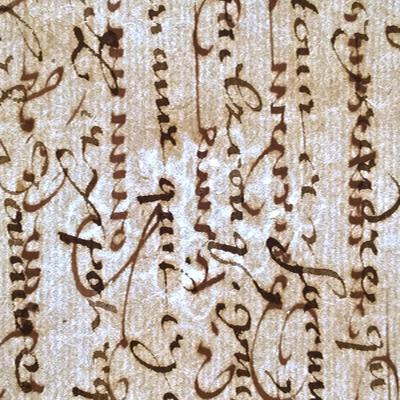}
		\includegraphics[width=0.075\linewidth]{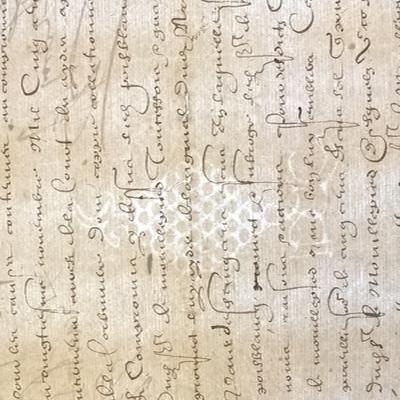}
		\includegraphics[width=0.075\linewidth]{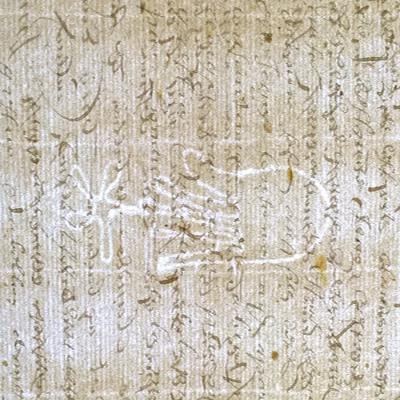}
		\includegraphics[width=0.075\linewidth]{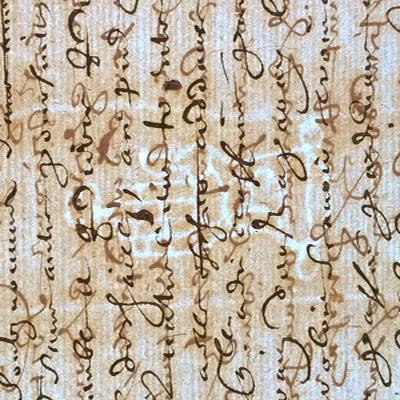}
	\end{center}
	\caption{Twelve examples (amongst 100) of triplets from our dataset A, with one clean reference and two normal photographs.}
	\label{fig:Atest}
 	\end{subfigure}
 	
	\begin{subfigure}{\linewidth}
		\begin{center}
			\includegraphics[width=0.075\linewidth]{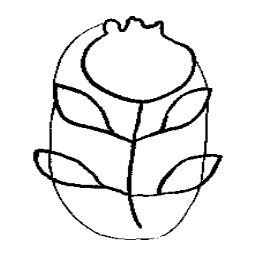}
			\includegraphics[width=0.075\linewidth]{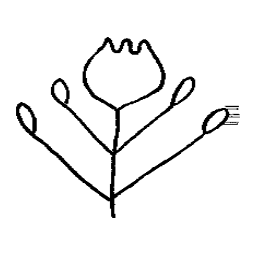}
			\includegraphics[width=0.075\linewidth]{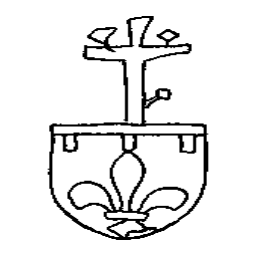}
			\includegraphics[width=0.075\linewidth]{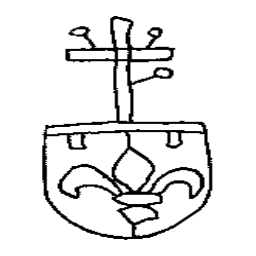}
			\includegraphics[width=0.075\linewidth]{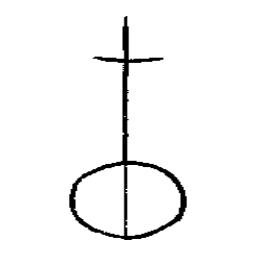}
			\includegraphics[width=0.075\linewidth]{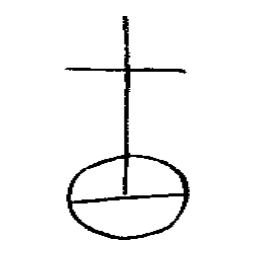}
			\includegraphics[width=0.075\linewidth]{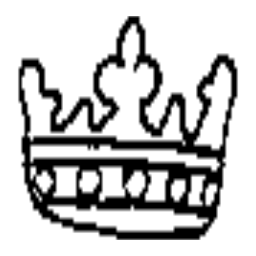}
			\includegraphics[width=0.075\linewidth]{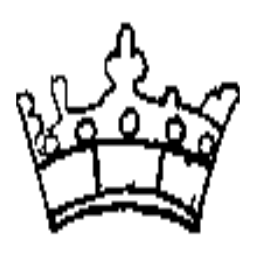}
			\includegraphics[width=0.075\linewidth]{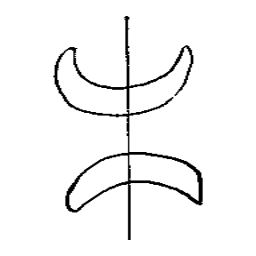}
			\includegraphics[width=0.075\linewidth]{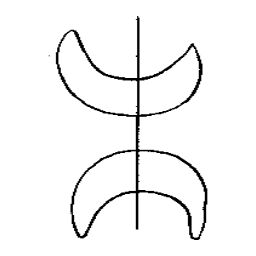}
			\includegraphics[width=0.075\linewidth]{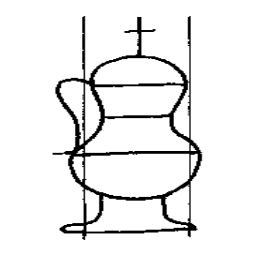}
			\includegraphics[width=0.075\linewidth]{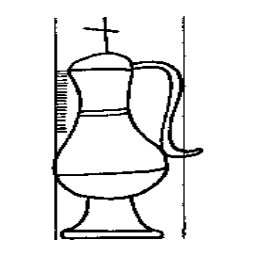}
			\\[1mm]
			\includegraphics[width=0.075\linewidth]{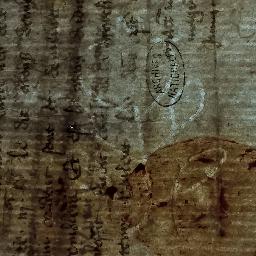}
			\includegraphics[width=0.075\linewidth]{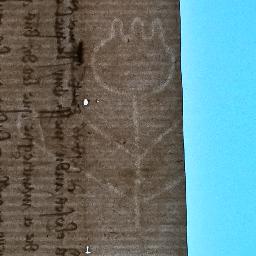}
			\includegraphics[width=0.075\linewidth]{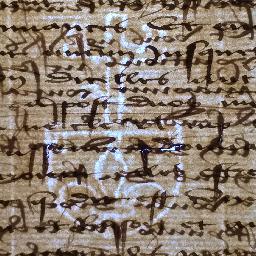}
			\includegraphics[width=0.075\linewidth]{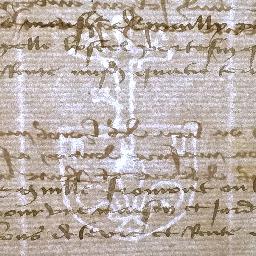}
			\includegraphics[width=0.075\linewidth]{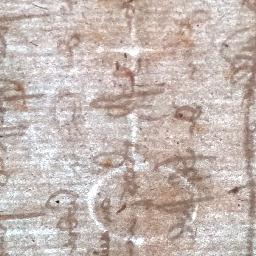}
			\includegraphics[width=0.075\linewidth]{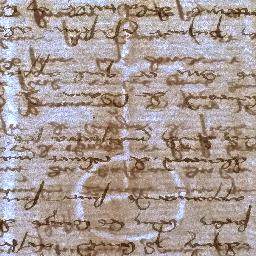}
			\includegraphics[width=0.075\linewidth]{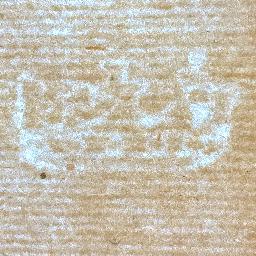}
			\includegraphics[width=0.075\linewidth]{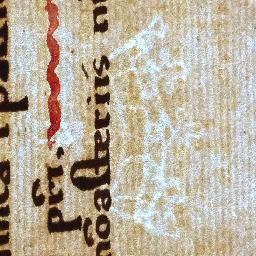}
			\includegraphics[width=0.075\linewidth]{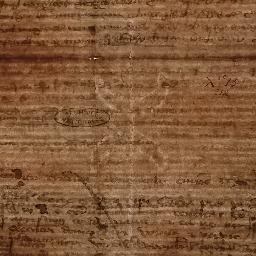}
			\includegraphics[width=0.075\linewidth]{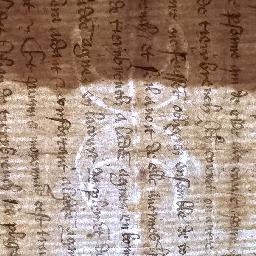}
			\includegraphics[width=0.075\linewidth]{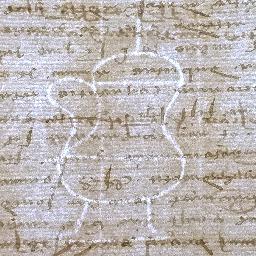}
			\includegraphics[width=0.075\linewidth]{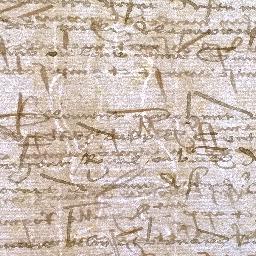}
			\\[1mm]
			\includegraphics[width=0.075\linewidth]{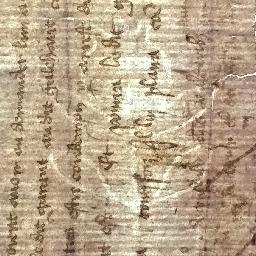}
			\includegraphics[width=0.075\linewidth]{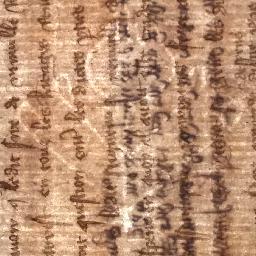}
			\includegraphics[width=0.075\linewidth]{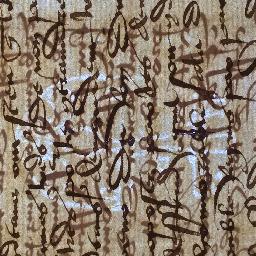}
			\includegraphics[width=0.075\linewidth]{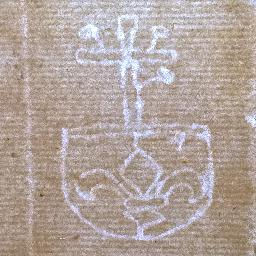}
			\includegraphics[width=0.075\linewidth]{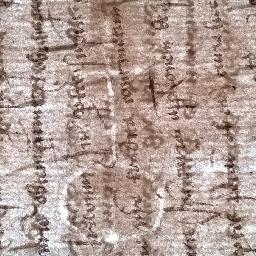}
			\includegraphics[width=0.075\linewidth]{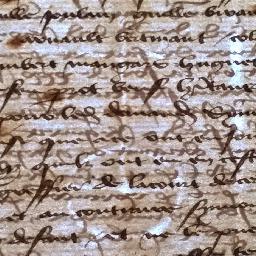}
			\includegraphics[width=0.075\linewidth]{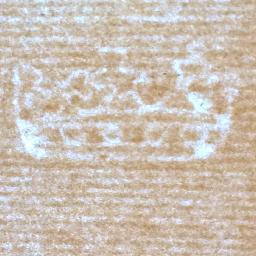}
			\includegraphics[width=0.075\linewidth]{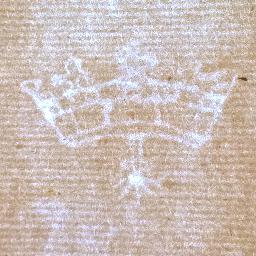}
			\includegraphics[width=0.075\linewidth]{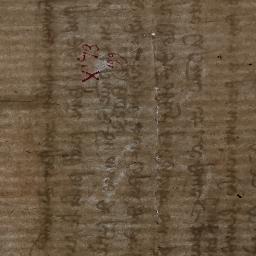}
			\includegraphics[width=0.075\linewidth]{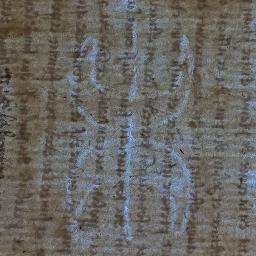}
			\includegraphics[width=0.075\linewidth]{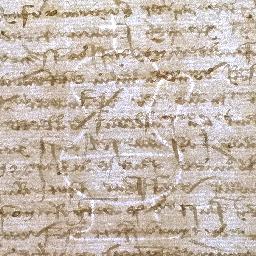}
			\includegraphics[width=0.075\linewidth]{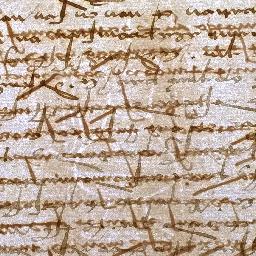}
		\end{center}
		\caption{Twelve examples (amongst 100) of test triplets from our dataset B, with one drawing reference and two normal photographs.}
		\label{fig:BTest}
	\end{subfigure}
 	
 	\begin{subfigure}{\linewidth}
 	\begin{center}
		\includegraphics[width=0.075\linewidth]{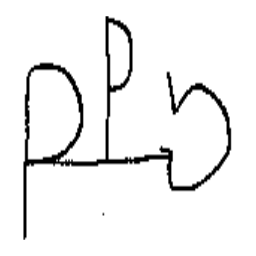}
		\includegraphics[width=0.075\linewidth]{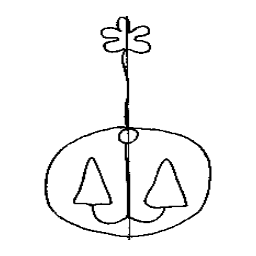}
		\includegraphics[width=0.075\linewidth]{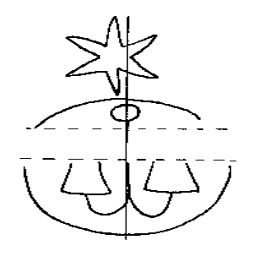}
		\includegraphics[width=0.075\linewidth]{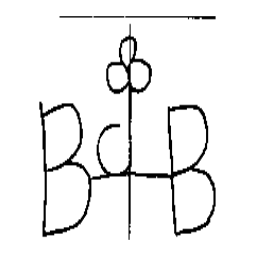}
		\includegraphics[width=0.075\linewidth]{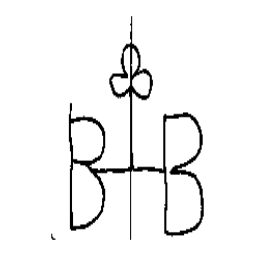}
		\includegraphics[width=0.075\linewidth]{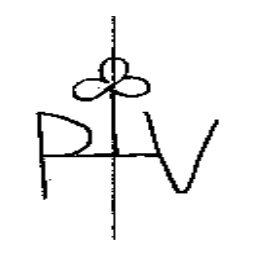}
		\includegraphics[width=0.075\linewidth]{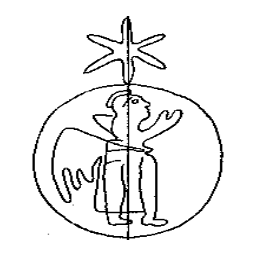}
		\includegraphics[width=0.075\linewidth]{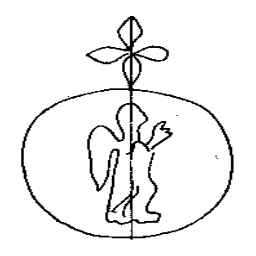}
		\includegraphics[width=0.075\linewidth]{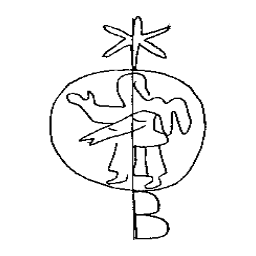}
		\includegraphics[width=0.075\linewidth]{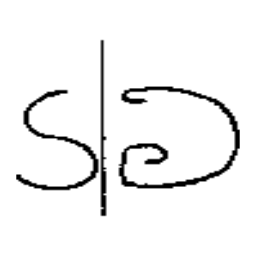}
		\includegraphics[width=0.075\linewidth]{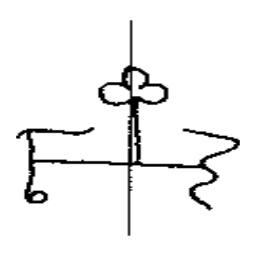}
		\includegraphics[width=0.075\linewidth]{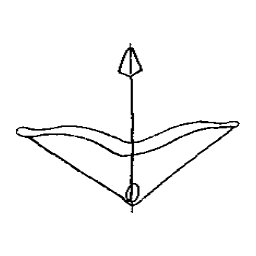}
		\vspace*{1mm}
		\includegraphics[width=0.075\linewidth]{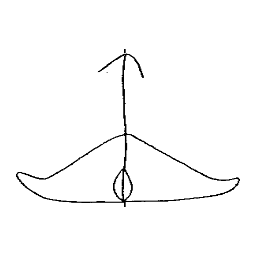}
		\includegraphics[width=0.075\linewidth]{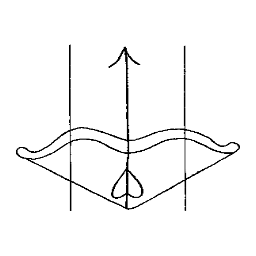}
		\includegraphics[width=0.075\linewidth]{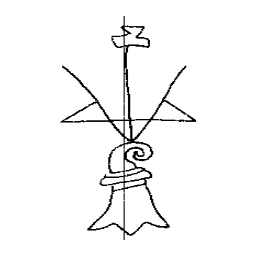}
		\includegraphics[width=0.075\linewidth]{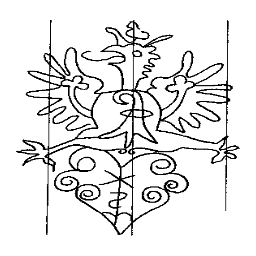}
		\includegraphics[width=0.075\linewidth]{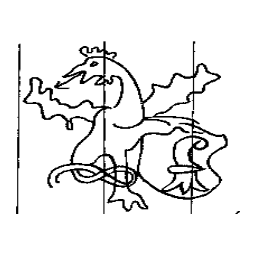}
		\includegraphics[width=0.075\linewidth]{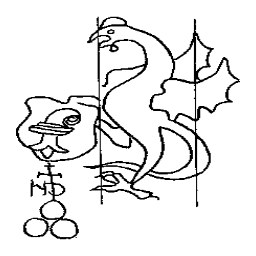}
		\includegraphics[width=0.075\linewidth]{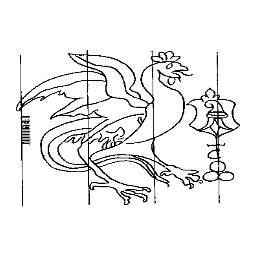}
		\includegraphics[width=0.075\linewidth]{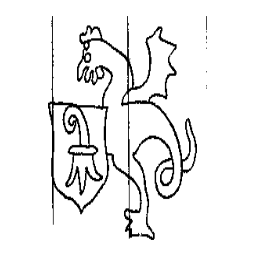}
		\includegraphics[width=0.075\linewidth]{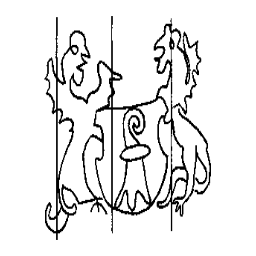}
		\includegraphics[width=0.075\linewidth]{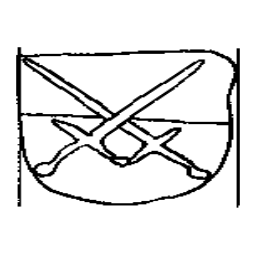}
		\includegraphics[width=0.075\linewidth]{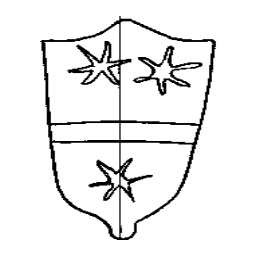}
		\includegraphics[width=0.075\linewidth]{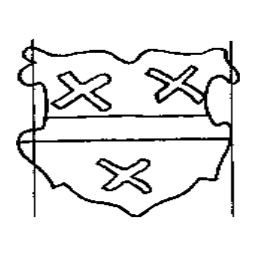}
		\vspace*{1mm}
		\includegraphics[width=0.075\linewidth]{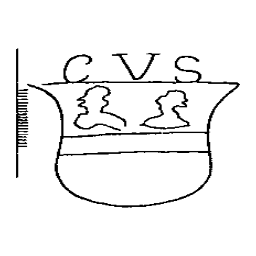}
		\includegraphics[width=0.075\linewidth]{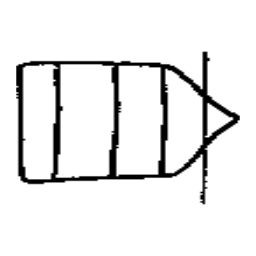}
		\includegraphics[width=0.075\linewidth]{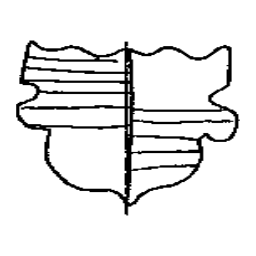}
		\includegraphics[width=0.075\linewidth]{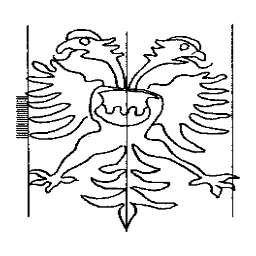}
		\includegraphics[width=0.075\linewidth]{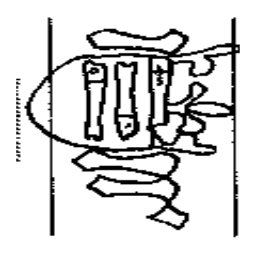}
		\includegraphics[width=0.075\linewidth]{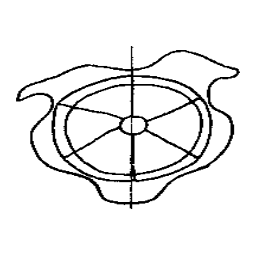}
		\includegraphics[width=0.075\linewidth]{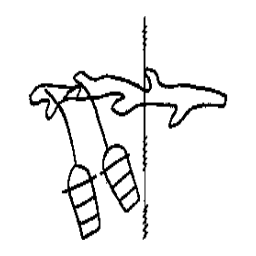}
		\includegraphics[width=0.075\linewidth]{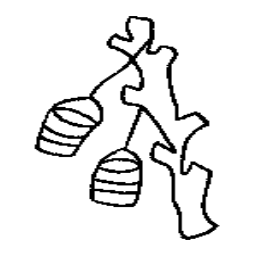}
		\includegraphics[width=0.075\linewidth]{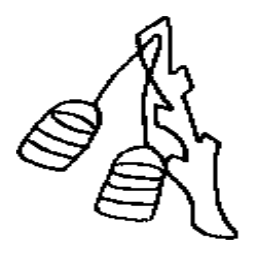}
		\includegraphics[width=0.075\linewidth]{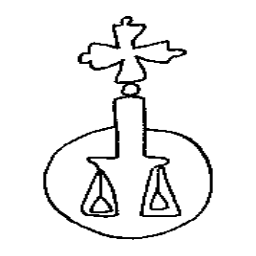}
		\includegraphics[width=0.075\linewidth]{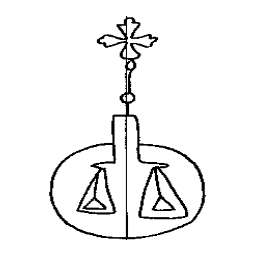}
		\includegraphics[width=0.075\linewidth]{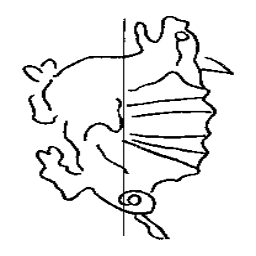}
		\vspace*{1mm}
		\includegraphics[width=0.075\linewidth]{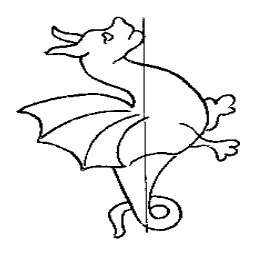}
		\includegraphics[width=0.075\linewidth]{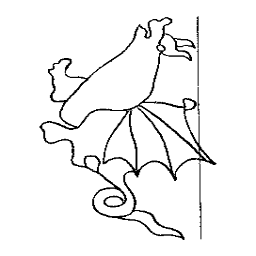}
		\includegraphics[width=0.075\linewidth]{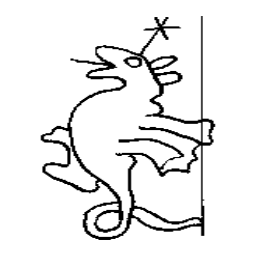}
		\includegraphics[width=0.075\linewidth]{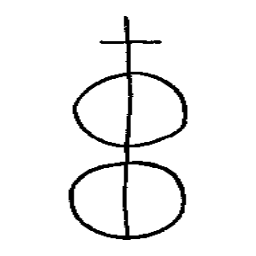}
		\includegraphics[width=0.075\linewidth]{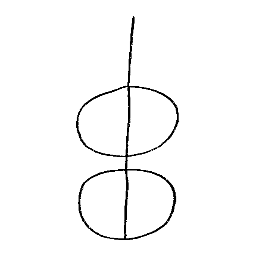}
		\includegraphics[width=0.075\linewidth]{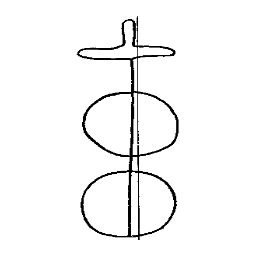}
		\includegraphics[width=0.075\linewidth]{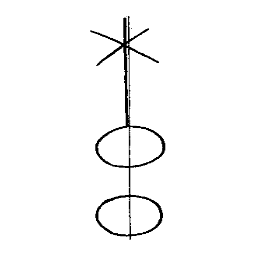}
		\includegraphics[width=0.075\linewidth]{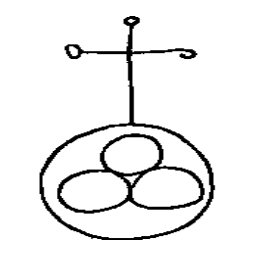}
		\includegraphics[width=0.075\linewidth]{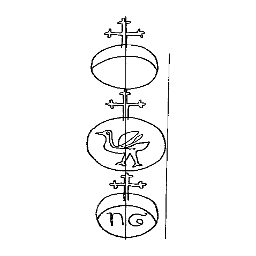}
		\includegraphics[width=0.075\linewidth]{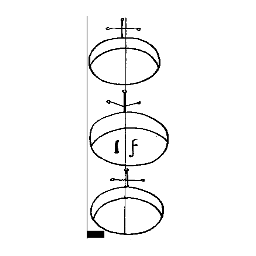}
		\includegraphics[width=0.075\linewidth]{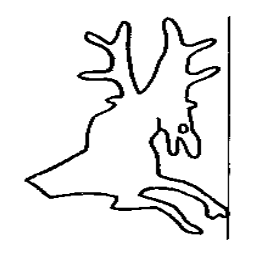}
		\includegraphics[width=0.075\linewidth]{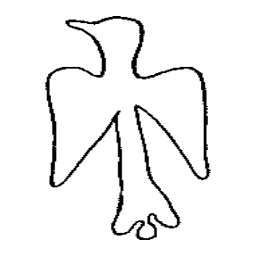}
	\end{center}
	\caption{Examples of the 16,753 drawings from our subset of the Briquet catalog. Notice the similarity between the classes.}
	\label{fig:BriquetM}
 	\end{subfigure}
 	
 	\caption{Examples of our datasets: (a) 36/60 photographs of the same watermark from a typical training category in our dataset A; (b) 12/100 test categories in dataset A, the first row is the clean reference photograph, the second and third rows are the two normal query photographs; (c)  12/100 categories in our dataset B, the first row is the reference drawing, the second and third rows are the two query photographs; (d) 48/16,753 different classes in our framed subset from the Briquet catalog. Note the diversity of appearance for the same watermark (a) and the presence of many very similar classes~(c,d)}
 	\label{fig:watermarkVis}
\end{figure*}

\IEEEPARstart{I}{n} this section, we explain how we created datasets to evaluate {\bf (A) one-shot } and {\bf (B) cross-domain} watermark recognition. The characteristics of our datasets are summarized in Table \ref{tab:data} and compared with existing datasets. Examples of images from our datasets are shown in Figure \ref{fig:watermarkVis}.\\

\paragraph*{Class definition} Some details of paper fabrication are important to understand the data variation and class definition. Watermarks were initially used by workshops to identify their production, and can thus be used to locate the origin of paper. They were created by small wires, shaped around a common model and added to the paper mold. Because molds were usually made in pairs (for reasons based on the typical paper-making workflow) and because the wire patterns became gradually deformed over time, the same pattern can be found with several small variations, which we still consider to be the same watermark. 

When the watermarks were replaced by new ones, formed around a new model, the new watermarks for the same workshop and the same sort of paper often had similar semantic content, but can be differentiated from the old ones, and we consider them as a different class. Indeed, being able to make this distinction is important to date the paper accurately.

To the best of our knowledge, we are the first ones to tackle at large scale the fine-grained recognition problem resulting from this definition, which is also the one from the main catalogs such as~\cite{briquet1907filigranes} and would be of strong practical interest.\\

\paragraph*{Photography and pre-processing procedure} Since our goal is to develop a procedure that could easily be applied without using any special or expensive device, we photographed watermarks with a standard cell phone camera (Microsoft Lumia 640 LTE). Since knowing the orientation of the watermark is impossible without interpreting its content, which is often challenging, we decided that the orientation would not be constrained.
The back-light was provided by a luminescent sheet. Since it is easy to visually identify the location of the watermark on a page, we required the photo to be taken so that the watermark would cover as much as possible of a rectangle with an aspect ratio of 2:3 visualized on the screen, with the largest dimension of the watermark contained within the largest dimension of the rectangle. We resize all images to transform the guiding rectangle into a square of size 224 and crop the surrounding 256 pixels wide square. \\

\paragraph*{Dataset A} The goal of this dataset is to train and evaluate methods for one-shot fine-grained watermark recognition from photographs. It thus only includes photographs and is split into two parts: a first part that can be used for feature/meta-training, with many examples of each watermark, and a second part, with few examples, to evaluate one-shot recognition. Obtaining a large-scale and diverse dataset of watermarks, with many examples for each fine-grained class, is very challenging. Our insight was to use watermarks found in notarial archives. Indeed notaries regularly bought paper in large quantities, and it was thus easier to collect many identical examples of many different watermarks. Because the leaves were left unbound, they are also easier to photograph. We first collected 50 training and 10 validation images for 100 watermarks, which we found was large enough to perform pre-training / meta-training of CNNs. An example of images from the same class is given Figure~\ref{fig:Atrain}. We then collected 3 photographs for 100 other test classes, one 'clean' image without any writing and two standard test images (see Figure \ref{fig:Atest}). We used the 'clean' images as references for one-shot recognition, as they are representative of what archivists typically collect as reference images and allow us to ensure that recognition is not related to the writing style of the document.
Note that many of the training and testing examples are very cluttered and in poor condition (e.g. Figure \ref{fig:teasera}), making the task very challenging.\\

\begin{figure}[t]
	\centering
	\begin{subfigure}{0.30\linewidth}
		\centering
		\includegraphics[width=\linewidth]{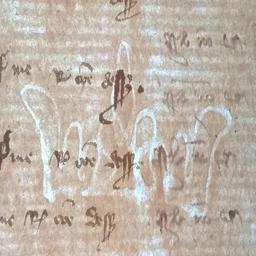}
		\caption{}
		\label{fig:query}
	\end{subfigure}~
	\begin{subfigure}{0.30\linewidth}
		\centering
		\includegraphics[width=\linewidth]{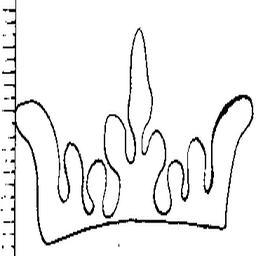}	
		\caption{}
		\label{fig:engraving}
	\end{subfigure}~
	\begin{subfigure}{0.30\linewidth}
		\centering
		\includegraphics[width=\linewidth]{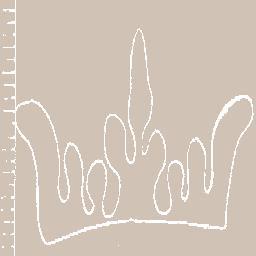}
		\caption{}
		\label{fig:synth}
	\end{subfigure}
	
	\begin{subfigure}{\linewidth}
		\centering
		\includegraphics[width=0.30\linewidth]{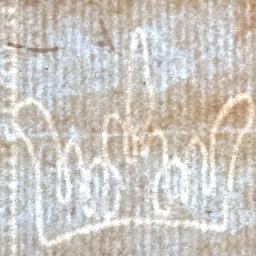}
		\includegraphics[width=0.30\linewidth]{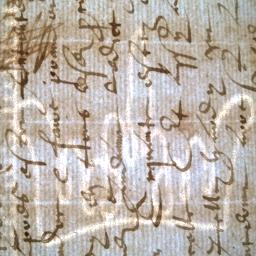}
		\includegraphics[width=0.30\linewidth]{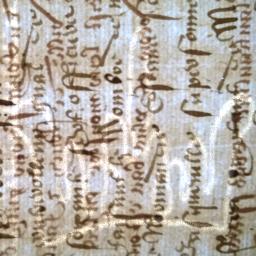}
		\caption{}
		\label{fig:randomization}
	\label{fig:transfer}
	\end{subfigure}
	\caption{Different types of images that we provide and generate for the classes of dataset B: (a) Photograph; (b) Drawing; (c) Synthetic Image; (d) Randomized synthetic images}
\end{figure}

\paragraph*{Ours Briquet} An important challenge is to use the drawings from existing watermark catalogs to perform recognition. Indeed, collecting watermarks and information for such catalogs is very tedious and expensive work, that cannot easily be reproduced. We focused on the Briquet catalog~\cite{briquet1907filigranes} whose drawings are entirely available online~\cite{briquetonline}. However, the available images include additional information, such as IDs of watermarks, paper line positions or complementary marks that can be found at another position on the paper sheets. While such information can be very valuable to experts, it cannot easily be used in a simple automatic recognition system based on a single photograph. We thus extracted the main part of the watermarks whenever it was clear and ended up with 16,753 drawings that could be used as reference for photograph recognition. Examples of these drawings can be seen in Figure~\ref{fig:BriquetM}. Notice that many watermarks are visually very similar to one another, making the classification task very challenging. Also notice that the drawings sometimes still include paper lines that are not part of the watermark but could not easily be removed. \\ 

\paragraph*{Dataset B} 
 The challenge that we want to address and evaluate with this dataset is to recognize photographs from watermarks based only on the drawing that we extracted from a catalog. We thus searched the original archives that provided the material for the Briquet catalog~\cite{briquet1907filigranes}, in a specific city (Paris). We were able to collect photographs for {240} classes, for which we also provide the published drawings (see examples in Figure \ref{fig:teaserc} and~\ref{fig:BTest}). Because comparing photographs (Fig.~\ref{fig:query}) directly with a line drawing (Fig.~\ref{fig:engraving}) as reference is very challenging, we also report results using a synthetic image generated from the drawing simply by using the average watermark color as background and making the drawing pattern lighter (Fig.~\ref{fig:synth}). Finally, we generated randomized synthetic images (Fig.~\ref{fig:randomization}) from the drawings. The randomized synthetic image $S$ are generated by computing $S=B+R \times (G*E) $, where $B$ is a background sampled from photographs of paper without watermarks, $G$ is a Gaussian filter, $R$ is a random image and $E$ is the binary watermark pattern extracted from the drawing. We split our {240} classes into {140} training and {100} validation classes. {In addition to the reference drawing, each of the validation classes includes two photographs and each of the training classes includes between one and seven photographs. The total number of photographs in the training set is 463.} This dataset allows to test cross-domain recognition, using drawings as reference, and recovering the class of a single test photograph. In our experiments, we first compare methods for 100-class one-shot cross-domain classification and then give results for the even more challenging 16,753-class classification.


\section{Local spatially-aware approach}
\label{sec:method}
\IEEEPARstart{I}{n} this section, we explain how we introduced geometric consistency in deep image matching and feature learning.

\subsection{Local matching similarity score}
\label{sec:matching}
We propose to compare images by computing mid-level CNN features on the test image and matching each of them densely in the reference images at {five} scales. For each local feature $f^i_1$, at position $x^i_1$ in the test image $I_1$ we consider its most similar feature $f^i_2$ at all pre-defined scales of the reference image $I_2$ and write the associated position $x^i_2$. We propose a local matching score which uses a combination of a 
{\it Spatial Consistency} score (SC) measuring the similarity between the positions $x^i_1$ and $x^i_2$ and a {\it Feature Similarity} (FS) measuring the distance between $f^i_1$ and  $f^i_2$:
\begin{eqnarray}
S(I_1, I_2) = \sum_{i \in \mathcal{I}} \underbrace{e^{-\dfrac{\|x^i_1-x^i_2\|^2}{2\sigma^2} }}_{SC} \underbrace{s(f^i_1,f^i_2)}_{FS}
\label{eqn:score}
\end{eqnarray}
 where  $S$ is the image level similarity we define, $s$ is a feature level similarity, for which we use cosine similarity in all of our experiments; $\mathcal{I}$ indexes the set of features in the test image and $\sigma$ is a tolerance parameter. 
We can directly use the difference in absolute position $\|x^i_1-x^i_2\|^2$ because we assume that the watermarks were coarsely aligned. If it were not the case, the above score could easily be used in conjunction with a RANSAC algorithm to identify candidate transformations, in a procedure similar to the discovery score proposed in ArtMiner~\cite{shen2019discovery}.

\begin{figure}[t]
	\centering
	\begin{subfigure}{0.35\linewidth}
		\centering
		\includegraphics[width=\linewidth]{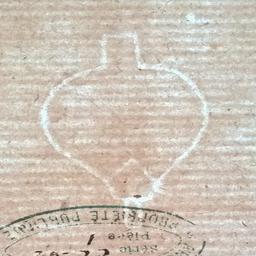}
		\label{fig:query_vis}
	\end{subfigure}~~~
	\begin{subfigure}{0.35\linewidth}
		\centering
		\includegraphics[width=\linewidth]{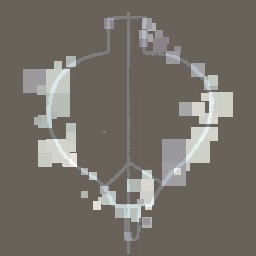}
		\label{fig:engraving_vis}
	\end{subfigure}
	\caption{Localization of the features contributing to our matching score (eq. \ref{eqn:score}). Left: query image; Right: reference image, where the brighter patches are the patches matched to the query}
	\label{fig:visual_correspondence}
\end{figure}

Note that for each feature $f^i_1$ only its best match in the reference image is considered. This implicitly removes any contribution for non-discriminative regions and for details that are only visible in one of the depictions, since the associated spatial-consistency part of the score (SC) will typically be zero. This is visualized in Figure~\ref{fig:visual_correspondence}, where the brighter patches in the right-hand image correspond to patches from the left-hand image that have been matched accurately enough to contribute to our similarity score, their brightness proportional to their contribution to the score. 

\subsection{Learning features for cross-domain recognition}
\label{sec:finetuning}
A key element in the matching score defined in the previous section, obviously, is the feature used for matching. The simplest approach is to use intermediate features from a network trained for watermark classification as local features. Here we aim at improving these features, in particular for cross-domain recognition.

We follow a metric learning approach. Assuming we have a set of positive pairs $\mathcal{P}$ and a set of negative pairs $\mathcal{N}$, we learn our features $f$ by minimizing a standard triplet loss: 
\begin{equation}
\begin{aligned}
\mathcal{L}(f) =\sum_{(n_1,n_2)\in \mathcal{N}} \max( 1-\lambda, \similarity(f(n_1),f(n_2)))   \\
-\sum_{(p_1,p_2)\in \mathcal{P}}\min(\lambda , \similarity(f(p_1),f(p_2))  
\label{eqn:triplet-loss}
\end{aligned}
\end{equation}
where the similarity measure $s$ is the cosine similarity and $\lambda=1.0$ in all our experiments.

The main challenge for learning such a feature is defining the sets of positive and negative pairs. A natural approach would be to consider that the local features of a photograph/drawing pair should match everywhere and that any other match is negative. However, the drawing and photograph might not be exactly aligned and there might be errors in the drawing. Additionally, an important proportion of the local features might not be discriminative. 
We thus propose an alternative extraction of positive and negative pairs, which depends on the current feature $f$. 
To define the positive pairs, we start from a local feature in a drawing and look for its best match in a photograph from the same watermark at {five} scales. We then keep the matching features as a positive pair only if the spatial distance between the center of the feature location is less than a threshold $\tau$. This allows to take into account misalignment between the source and target images. It also gives robustness to errors and differences, such as the small lines at the left of figure \ref{fig:engraving} which are not clearly visible in the photograph. Once a positive pair has been identified, we look for hard negatives by matching the source feature to all the photographs of other watermarks and select the most similar feature as negative.

As for our matching score, the specificity of this training procedure is that it is performed at the level of local features, not at the image level and that it is designed to explicitly take into account spatial misalignment and differences. We demonstrate the benefits of this approach in Section \ref{sec:exp_cd}.

\subsection{Implementation details}
For all our experiments, we use a ResNet-18~\cite{he2016deep} architecture, larger networks leading to either similar or worse performances. Since the photographs might be flipped and rotated with respect to the references, we consider matches with four rotated (0$^\circ$, 90$^\circ$, 180$^\circ$, 270$^\circ$) reference images and their flipped versions in all the experiments.  Both  for local matching and feature fine-tuning, we used {\it conv4} features. Each source image was resized to 352$\times$352 so that it was represented by 22$\times$22 features. To be robust to scale discrepancies, we matched the source features to features extracted from the target image resized at {five} scales, corresponding to 16, 19, 22, 25 and 28 features. Our models were trained with the Adam~\cite{kingma2014adam} optimizer with  $\beta = [0.9, 0.99]$ and a learning rate of 1e-3 for feature initialization and 1e-5 for fine tuning. Using a single GPU Geforce GTX 1080 Ti, pre-training for classification converged in approximately 5 hours and fine-tuning on approximately 2 hours, and matching a pair of images took approximately 37ms. 

\section{Experiments}
\label{sec:results}
\label{sec:ResultsDatasetA}
\IEEEPARstart{I}{n} this section, we present our results and compare with baselines first for one-shot recognition then for cross-domain one-shot recognition. Finally, we provide a qualitative analysis of our results. 

For both tasks, we pre-trained a network for classification on the 100 meta classes of dataset A with 60 images per class, using 50 for training and 10 for validation. We obtained the best performances by training with a strong dropout (0.7 ratio), which is not surprising given the relatively small size of our dataset. We obtained a validation accuracy of $98.8\%$, the mis-classified images being only very difficult or ambiguous cases, which shows that our 6k images dataset was large enough to train a good network for fine-grained watermark classification. Note that we also tested features trained for ImageNet classification~\cite{deng2009imagenet} as well as the SIFT features using geometric verification~\cite{lowe2004distinctive,sivic2003video}, but both lead to performances close to chance because the features were not adapted to the task.

{For all baseline approaches, images are firstly resized to $256 \times 256$ then cropped the central $224 \times 224$ region, which is the same pre-processing procedure as in the classification on dataset A. We report the best performance over {\it conv4}/{\it conv5} features. The optimal parameters and training strategy for each baseline approach are reported in the supplementary material\footnote{ \url{http://imagine.enpc.fr/~shenx/Watermark/supp.pdf}}}.

\subsection{One-shot recognition.}

In this section, we evaluate our local matching score (eq. \ref{eqn:score}) for one-shot recognition. We first compare our local matching score with state-of-the-art one-shot recognition approaches on dataset A for which the domain gap is limited and standard one-shot recognition approaches can be expected to work (Table~\ref{tab:oneshotresults}). We then compare it with other feature similarities on the more challenging dataset B (Table~\ref{tab:ablation_study_match}).  

\begin{table}[t!]
	\centering
	\begin{tabular}{|l|c|c|c|}
		\hline    
		Method $\backslash$ Features &  AvgPool & Concat & \makecell{Local Sim.}  \\ \hline
		Baseline & 69 & 74 & 75 \\
		Cosine Classifier~\cite{qi2018low,gidaris2018dynamic} &  84 & 82 & 84 \\
		Matching Networks (scratch)~\cite{vinyals2016matching}  &  73 & 76 & 80\\
		Matching Networks (ft)~\cite{vinyals2016matching} & 74 & 76 & 82\\
		Weights Prediction~\cite{gidaris2018dynamic} & 86 & 84 &  85 \\ \hline
		Ours (Resolution 256) & \multicolumn{3}{c|}{85} \\ 
		Ours (Resolution 352) & \multicolumn{3}{c|}{\bf 90} \\
		\hline
		
	\end{tabular}
	\vspace{2mm}
	\caption{ Comparison with state of the art one-shot recognition approaches on dataset A (200 images to compare to classify in 100 categories unseen during training and described by a single 'clean' image). Accuracy in \%. Our score based on local matches clearly outperforms all baselines.}
	\label{tab:oneshotresults}
\end{table}

\begin{table}[t!]
	\centering
	\begin{tabular}{l|c||c|c|c}
		    
		Method &  \makecell{A} & \makecell{B-\ref{fig:engraving}} & \makecell{B-\ref{fig:synth}} & \makecell{Time (s / Query)}  \\ \hline
		\multicolumn{5}{c}{\bf Exact features comparison} \\
		AvgPool  &  69 & 4 & 12 & 1\\
		Concat  & 74 & 55 & 61 & 2\\
		Local Sim. & 75 & 56 & 65 & 2\\
		Discovery ~\cite{shen2019discovery} & 88 & 51 & 63 & 420\\
		
		\multicolumn{5}{c}{\bf  Our local matching score} \\
		Ours &  \bf 90 & \bf 66 & \bf 72 &  15\\
		
		\hline
		
	\end{tabular}
	\caption{Comparison of {our local} matching score (eq. \ref{eqn:score}) with alternative feature similarities. We report percentage of accuracy for one-shot recognition on dataset A (A column) and on dataset B using either the drawing (B-\ref{fig:engraving}) or our synthetic image (B-\ref{fig:synth}). }

	\label{tab:ablation_study_match}
\end{table}

\paragraph*{Comparison to state-of-the-art one-shot recognition methods}

On dataset A, which does not include any domain shift, we compare our method to some recent one-shot recognition approaches :
\begin{itemize}

    \item \emph{Baseline} : directly using the features learned by training a network on the classification task. 
	
	\item  \emph{Cosine Classifier} : recent work~\cite{gidaris2018dynamic,qi2018low} has shown that the performance of the baseline can be improved if during training the dot-product operation (between classification weights and features) in the last linear layer of the network is replaced with the cosine similarity operation.
	
	\item  \emph{Matching Networks}: we tried the metric-learning approach of Matching Networks~\cite{vinyals2016matching}, performing meta-training to solve one-shot recognition tasks using a differentiable nearest-neighbor-like classifier. We tested either directly training it from random initialization ("scratch" in Table~\ref{tab:oneshotresults}) or fine-tuning it from a network pre-trained on the classification task of dataset A ("ft" in Table~\ref{tab:oneshotresults}). In the latter case the pre-trained network uses a cosine-similarity based classifier.
	
    \item \emph{Weights Prediction}: the one-shot recognition approach of Gidaris and Komodakis~\cite{gidaris2018dynamic} predicts classification weights used by the last linear layer of a cosine-similarity-based network from a single training example of the category. It uses a feature extractor learned with a cosine-similarity based classifier which remains frozen during the meta-training procedure.
\end{itemize}

For each feature, we report three different similarities:
\begin{itemize}
    \item {\it AvgPool} : cosine similarity using the average pooled features.
    \item {\it Concat} : cosine similarity on the descriptor formed by the concatenation of  all the spatial features.
    \item {\it Local Sim.} : computing the cosine similarity over each local feature individually, then averaging.
\end{itemize}
Note that cosine similarity consistently performs better than dot product.

{As can be seen in Table~\ref{tab:oneshotresults}, our matching score leads to 85\% accuracy which is close to the best one-shot approach, Weights Predictions~\cite{gidaris2018dynamic}, but without any specific feature learning. This demonstrates the interest of our local matching score for one-shot fine-grained watermark recognition. The performance can further be boosted to 90\% by resizing image to a larger resolution, $352\times 352$ pixels, which we use in the rest of the paper.}\\


\paragraph*{Comparison of feature similarities for one-shot cross-domain recognition}
\label{sec:ablation}

In Table~\ref{tab:ablation_study_match}, we compare our {local matching} score (eq. \ref{eqn:score}) with alternative feature similarities on our two datasets. On dataset B, we use either the drawing (B-\ref{fig:engraving})  or our synthetic image (B-\ref{fig:synth}) as reference.  We always used the features trained for classification on dataset A, and compare on each dataset to the similarities described in the previous paragraph (AvgPool, Concat and LocalSim.) and to the discovery score proposed in our previous work ArtMiner~\cite{shen2019discovery}. Similar to our local matching score, the discovery score from ArtMiner considers the spatial location of the matches but relies on RANSAC to fit an affine transformation model. 
Our local matching score consistently outperforms all these baseline similarities. Interestingly, the discovery score~\cite{shen2019discovery} works better than other baselines and very similarly to ours on dataset A but is clearly worse than ours on dataset B. We think the reason is that the features are good enough to estimate good transformations on dataset A but fail on dataset B.    In our naive implementation, our approach is slower than direct feature comparison, but both can be mixed to obtain fast results on very large datasets (see large scale experiments in section~\ref{sec:exp_cd}).\\

\subsection{Cross-domain recognition.}
\label{sec:exp_cd}
\begin{table}[t!]
	\centering
	\begin{tabular}{|l|c|c|}
		\hline
		Method & B-\ref{fig:engraving} & B-\ref{fig:synth}\\ \hline
		\multicolumn{3}{|c|}{\bf Baselines} \\
		w/o Fine-tuning & 66 & 72 \\
        
		Unsupervised (Translation) & 63 &  70 \\ 
		Supervised (Affine) &64 &72 \\
		Randomization &53 & 75 \\
		Triplet-loss&  64 & 65\\
		NC-Net~\cite{Rocco18b} & 61 & 65  \\
		ArtMiner~\cite{shen2019discovery}&  60 & 71\\
		\multicolumn{3}{|c|}{\bf Our spatially-aware fine-tuning} \\ 
		$\tau =0$& 65 &  72\\ 
		$\tau =3/22$& \bf 75 & \bf 83  \\ 
		$\tau =\inf$ & 61 &  74  \\

		\hline 
	\end{tabular}

	\caption{Accuracy (in \%) on one-shot cross-domain recognition with different methods and different reference images ("B-\ref{fig:synth}" referred to our synthetic image and "B-\ref{fig:engraving}" referred to drawing). Standard domain adaptation strategies provide little improvement when combined with our score, while the fine-tuning described in section \ref{sec:finetuning} provides a clear boost.}

	\label{tab:ablation_study_finetune}
\end{table}

We now focus on cross-domain recognition. We first compare our approach with different feature training strategies for one-shot cross-domain 100-class recognition on our dataset B. We then demonstrate the effectiveness and generality of our approach by evaluating it on a fine-grained Sketch-Based Image Retrieval dataset. Finally, we focus on scaling our watermark recognition method to the full Briquet catalog, showing we can perform classification with more than 16k classes.\\

\paragraph*{Feature training}  In Table~\ref{tab:ablation_study_finetune}, we compare the results from our fine-tuning strategy to different baselines:
\begin{itemize}
    \item {\it Unsupervised (translation)}: in a spirit similar to ~\cite{sun2016return}, we translated the features in our target domain so that, on our training set, they have the same mean as the features from the target domain. We then use our score to perform nearest-neighbor classification.
	
	\item {\it Supervised (affine)}: since we have aligned images from both domains, we can adapt features from the source and target domains in a supervised way, similarly to \cite{massa2016deep,rad2018feature}. We found that a simple affine adaptation gave the best results, likely because of the small size of our dataset.
	
	\item {\it Randomization}: we trained a standard classifier using random images  such as the ones presented in Figure~\ref{fig:randomization} and generated as described in \ref{sec:datasets}. Such an approach has been shown to be very successful for example for 3D pose estimation \cite{su2015render,Loing2018}.
	
	\item {\it Triplet-loss}: similarly to our method, we tried to improve the features using a triplet loss on local features, but using as positive all aligned features in the images from the same category. 
    
    \item {\it NC-Net}~\cite{Rocco18b}: while it was not initially designed for domain adaptation, we trained NC-Net on our database because of the intuition that, similarly to our method, it is able to learn to leverage spatial information. We use our pre-trained ResNet for the feature extractor and freeze it during the training. The other parts are kept the same as the category level matching model proposed in~\cite{Rocco18b}. The positive pairs are composed with one image from each domain, which results in 463 pairs in the training set. The training converges in 20 epochs. We then consider the sum of the scores over all correspondences as the score between a pair of images.
    
    \item {\it ArtMiner}~\cite{shen2019discovery}: the two main differences with our approach is that ArtMiner does not use the category supervision and the approximate alignment. Instead, it leverages spatial consistency in the matches to select positive and negative training pairs.
\end{itemize}

All results except for NC-Net are reported with our local matching score as using different scores (including the classification score for domain randomization) 
leads to worse performances. This might be the reason why standard domain adaptation approaches only marginally improve performances over the baseline. Another possible reason is the small size of our training set (463 photographs, 140 references) for which these methods might not be adapted.
On the contrary our fine-tuning strategy, which is explicitly designed to handle clutter, misalignment and errors, boosts performances by a clear margin. In Table~\ref{tab:ablation_study_finetune} we report our results with no and infinite misalignment tolerance $\tau$. We provide a full analysis of the dependency of our results on $\tau$ {in the supplementary material\footnote{ \url{http://imagine.enpc.fr/~shenx/Watermark/supp.pdf}}}.\\ 


 


\paragraph*{Fine-grained Sketch-Based Image Retrieval (SBIR)} To demonstrate the generality of our approach we evaluated it on the SBIR task. We report results on the dataset of \cite{yu2016sketch} which consists of sketch-photo pairs of shoes and chairs. The shoes dataset contains 304 sketch-photo pairs for training and 115 for testing and the chairs dataset 200 pairs for training and 97 pairs for testing. We use our local {matching} score with the {\it conv4} features of a ResNet-18~\cite{he2016deep} architecture pre-trained on ImageNet~\cite{deng2009imagenet} and compare results with and without fine-tuning. We find that large image resolution and spatial tolerance $\sigma$ (see equation \ref{eqn:score}) achieve better performance. We use images of 384x384 pixels, corresponding to 24x24 feature maps in the {\it conv4} layer and $\sigma$ corresponding to 4 features. The results are reported in Table~\ref{tab:sbir}. Note that our local matching score alone allows to obtain performances above state of the art. Our fine-tuning strategy further provides a significant boost compared to ImageNet pre-trained weights. Our top-1 accuracy with fine-tuned model outperforms other methods by a significant margin on both datasets, demonstrating the interest of applying our approach for solving cross-domain recognition tasks beyond watermark recognition. \\

\begin{table}[t!]
		\centering
		\scriptsize
		\begin{tabular}{lcccc}
			\hline
			 \thead{Method}  & \multicolumn{2}{c}{Shoes} & \multicolumn{2}{c}{Chairs} \\
			
			 \thead{} & \thead{acc.@1} & \thead{acc.@10} & \thead{acc.@1} & \thead{acc.@10} \\
			\hline
			GDH@128bits~\cite{zhang2018generative} & 35.7 & 84.3 & 67.1 & 99.0 \\ 
			TSN~\cite{yu2016sketch} &  39.1 & \bf 87.8 & 69.1 & 97.9\\ 
			EdgeMAC~\cite{radenovic2018deep}& - & - &  85.6 & 97.9 \\
			
			\hline
			  
			 Ours with ImageNet feature & 40.0 & 85.2 &  90.7 & 99.0\\
			 Ours with Fine-tuned feature & \bf 52.2  & \bf 87.8 &  \bf 91.8 &\bf 1.0  \\  
			
			\hline
		\end{tabular}
		\caption{Results on fine-grained sketch-based
image retrieval (dataset V1 in \cite{yu2016sketch}). Our local matching score with ImageNet trained features provide results similar to the state of the art, and our fine-tuning provides an additional boost.}

		\label{tab:sbir}
\end{table}

\paragraph*{Large-scale recognition} We finally evaluate one-shot cross-domain recognition using the test photographs of our dataset B and our full curated version of the Briquet dataset as reference. This recognition with 16,753 fine-grained classes is very challenging, but also corresponds to a realistic scenario for watermark recognition. We use our synthetic images (Figure~\ref{fig:synth}) to represent the drawings. Since our local matching score is computationally expensive, we apply a two-step procedure for recognition. For each test photograph, we first select the top-$N$ candidate classes using direct feature comparisons and re-rank them using our local matching score. Table~\ref{tab:Briquet_results} shows the top-1 and top-1000 accuracy using the different baseline similarities described in Section \ref{sec:ablation}. Using the local similarity, i.e. averaging the cosine distance between the local features over the images leads to the best results, 28\% top-1 and 83\% top-1000 accuracy. We thus use it to perform the first step of selection. Re-ranking the top 1000 matches with our local similarity score boosts the top-1 accuracy to 54\%. {The best performance we can achieve is 55\% by ranking all the matches with our local matching score (N = inf).

In practice, selecting a small number of candidates is important to keep computational time low, and a user could easily look at the top-K results. We thus studied in detail the dependency of our results with respect to $N$ and $K$ and report the results in Figure~\ref{fig:briquet}. We can observe several interesting facts. First, fine-tuning clearly improves performance when using our local matching scores, but mostly degrades performance for local similarity. Second, results with our local matching scores are always clearly superior and increase with N. Third, accuracy for our local matching score increases very quickly for the top-10 matches, suggesting that a large proportion of our failures are actually due to ambiguous cases with very similar watermarks, which is verified qualitatively (c.f. Figure~\ref{fig:topRetriveal} and full results on the project website). 


In terms of computational time, it takes approximately 3s to rank the reference drawings with Local Similarity and {37}s to re-rank the top-1000 with our local matching score on a single GPU Geforce GTX 1080 T. This is acceptable for practical applications of our algorithm, and leads to 70\% top-10 accuracy. We thus believe the application of our algorithm will be a game-changer and widen considerably the potential use of watermark analysis, which until now has been limited to a small number of experts. 

\begin{table}[t!]
		\centering
		\scriptsize
		\begin{tabular}{lcccc}
			\hline
			 \thead{Method}  & \multicolumn{2}{c}{Briquet-\ref{fig:synth}} & \multicolumn{2}{c}{Briquet-\ref{fig:synth}+Fine-tuning} \\
			
			 \thead{} & \thead{acc.@1} & \thead{acc.@1000} & \thead{acc.@1} & \thead{acc.@1000} \\
			\hline
			AvgPool  &  0&  16&   0& 21\\
		    Concat  &   27& 77&    29&  82\\
		    Local Sim. & 28 & 80  & 28 & 83 \\
			
    		Ours N=1000&  \bf45 & 80 & 54 & 83  \\
			Ours N=inf&  44 & \bf86 & \bf 55 & \bf 91 \\
			
			\hline
		\end{tabular}
		\caption{Top-1 and top-1000 accuracy on our Briquet dataset with different models ("Briquet-\ref{fig:synth}" referred to using model trained on classification on dataset A and "Briquet-\ref{fig:synth}+Fine-tuning" referred to using our fine-tuned model): the approaches are AvgPool, Concat, Local Similarity and first applying Local Similarity to obtain N = 1000 top ranked references then using our score to re-rank the N references. }

		\label{tab:Briquet_results}
\end{table}

\begin{figure}[t]
	\centering
 	\includegraphics[width=\linewidth]{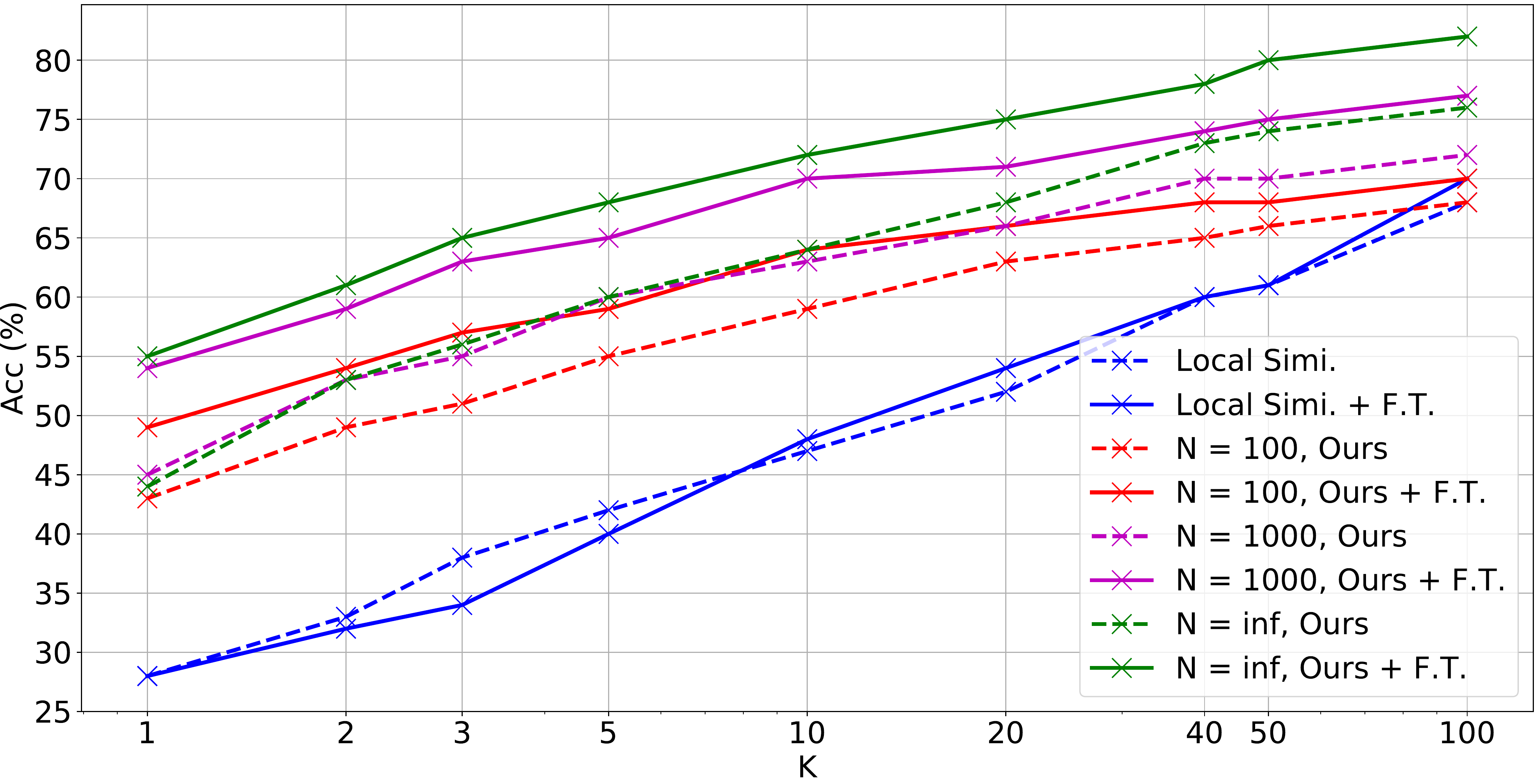}
 	\caption{Top-K accuracy on Briquet dataset. We first select the top-N candidates using Local Similarity (blue lines) then re-rank them with our score (other lines). The dashed lines correspond to features trained on our dataset A and the solid lines correspond to our fine-tuned model.}
 	\label{fig:briquet}
 	\vspace{-4mm}
 	
\end{figure}

\subsection{Qualitative analysis}

\label{sec:qualitative}

\begin{figure*}[t]
	\centering
 	\begin{subfigure}{\linewidth}
 	\includegraphics[width=\linewidth]{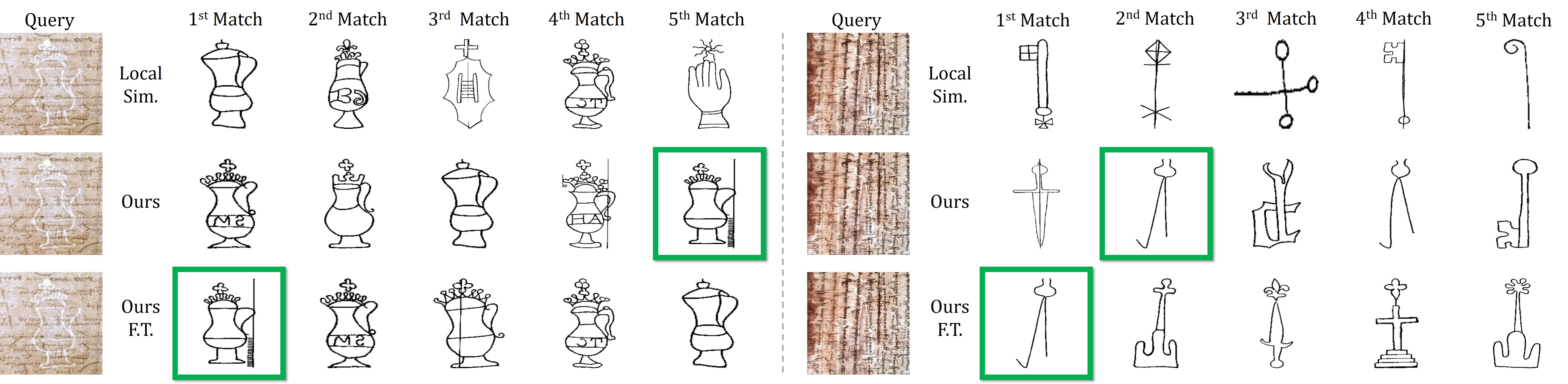}
 	\caption{Identifying a watermark from a photograph directly using the drawings~\ref{fig:engraving}. Note the many similar watermarks on the left.}
	\label{fig:plotEngraving}
 	\end{subfigure}
 
 	\begin{subfigure}{\linewidth}
 	\includegraphics[width=\linewidth]{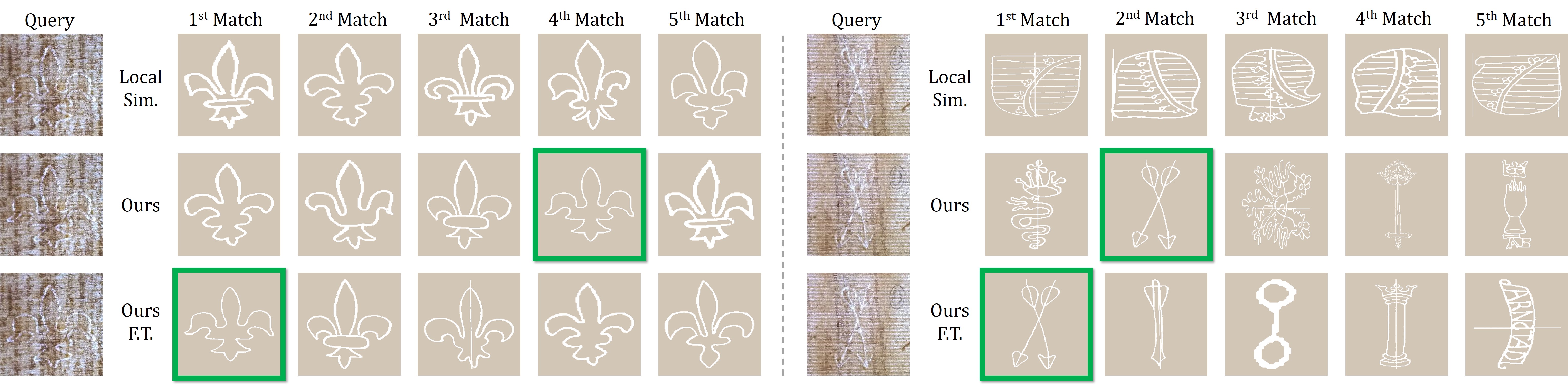}
 	\label{fig:plotSynthetic}
 	\caption{Identifying a watermark from a photograph using our synthetic images~\ref{fig:synth}. Note the many similar watermarks on the left.}
 	\end{subfigure}
 
 	\begin{subfigure}{\linewidth}
 	\includegraphics[width=\linewidth]{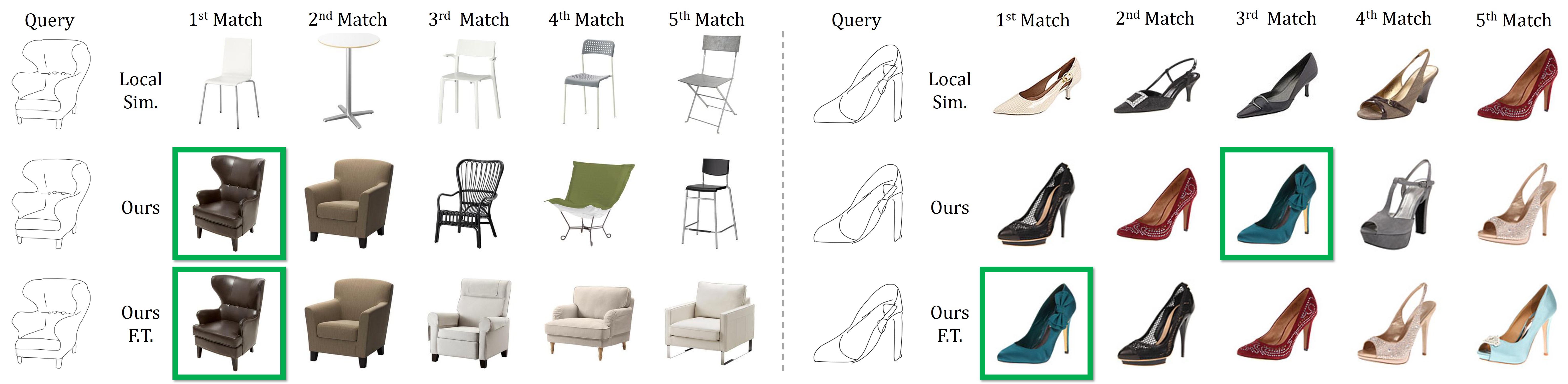}
 	\caption{Photograph retrieval from a drawing.}
	\label{fig:plotShoes}
 	\end{subfigure}
 	\caption{Example of top-5 matches using local similarity {(denoted as 'Local Sim.')} and our local matching score with and without feature fine-tuning {(denoted as 'Ours F.T.' and 'Ours' respectively)} on the different types of data we use for cross-domain one-shot retrieval.}
 	\label{fig:topRetriveal}
 	\vspace{-4mm}
\end{figure*}

In this section, we provide a qualitative analysis of our results. We first focus on our local matching score, then outline the effect of fine-tuning and finally discuss the quality of our results and failure modes. More visual results can be found on our project website \footnote{\url{http://imagine.enpc.fr/~shenx/Watermark}}. 

 Representative examples of top-5 matches on the different type of data we experimented with are shown in Figure~\ref{fig:topRetriveal} for the best similarity baseline (Local Similarity) and our local matching score before and after fine-tuning. We identified coarsely two types of watermarks, illustrated respectively on the left and right of Figure~\ref{fig:topRetriveal}: (i) watermarks that correspond to a common type, with many very similar classes (ii) watermarks which are more unique. \\

\paragraph*{Local matching score} 
The advantage of our local matching score for watermarks is clearest in case (i) where many similar classes exist. Indeed, the baseline local similarity typically gives very similar scores to all similar watermarks, while our local matching score ranks the exact match much better. 
To understand why, we visualize in Figure~\ref{fig:part1Vis} the contribution of the different parts of the watermark to different similarity scores: the AvgPool, Concat and Local similarity baselines described in Section \ref{sec:ablation} and our score. More precisely,  we compute the contribution of each local feature to the total score, and show the percentage of contribution of each feature on the watermark contour. Warmer colors correspond to higher contributions, the scale is the same in all images. 
In all the baseline approaches, the dominant light blue color corresponds to the fact that many regions have a small but non-negligible contribution to the final score. On the contrary with our local matching all regions that are not discriminative or accurate enough have no contribution (in dark blue) and regions that can be matched accurately have a similarly high contribution (red). For example, note on the first line of Figure~\ref{fig:part1Vis} that the paper line on the right of the drawing, that is not present in the actual watermark, has a non-negligible contribution for all similarity scores except ours. We believe this is the main reason for the superiority of our local matching score over baselines.

\begin{figure}[t]
	\centering
 	\includegraphics[width=0.95\linewidth]{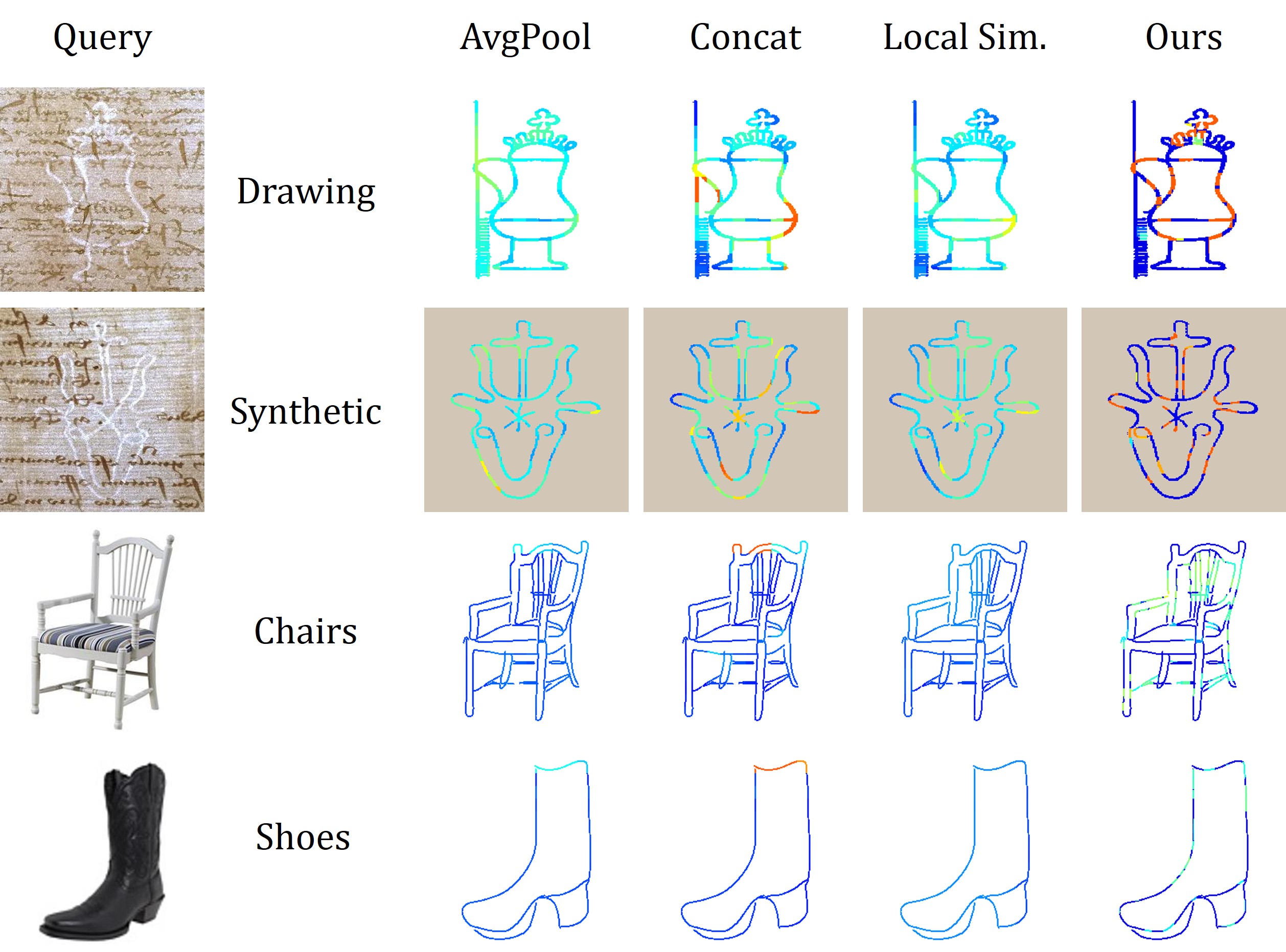}
 	
 	\caption{Visualization of the contribution of each feature to the similarity on our different datasets. The color map is set to blue if the contribution is 0 and red if the contribution is larger than 1\% of the similarity score. Note that our score only values specific regions corresponding to features that can be matched reliably.}
 	\label{fig:part1Vis}
 	\vspace{-4mm}
\end{figure}

\paragraph*{Effect of fine-tuning} Qualitatively, fine-tuning does not completely change the top matches but again improves the rank of exact matches, as in the examples of Figure~\ref{fig:topRetriveal}. To outline the effect of our fine-tuning we show in Figure~\ref{fig:part2Vis} a visualization of the contribution of each feature similar to the one described in the previous paragraph, comparing our score before and after fine-tuning in all four datasets. After fine-tuning, more discriminative regions have been matched and their scores become larger.


\begin{figure}[t]
	\centering
 	\includegraphics[width=0.8\linewidth]{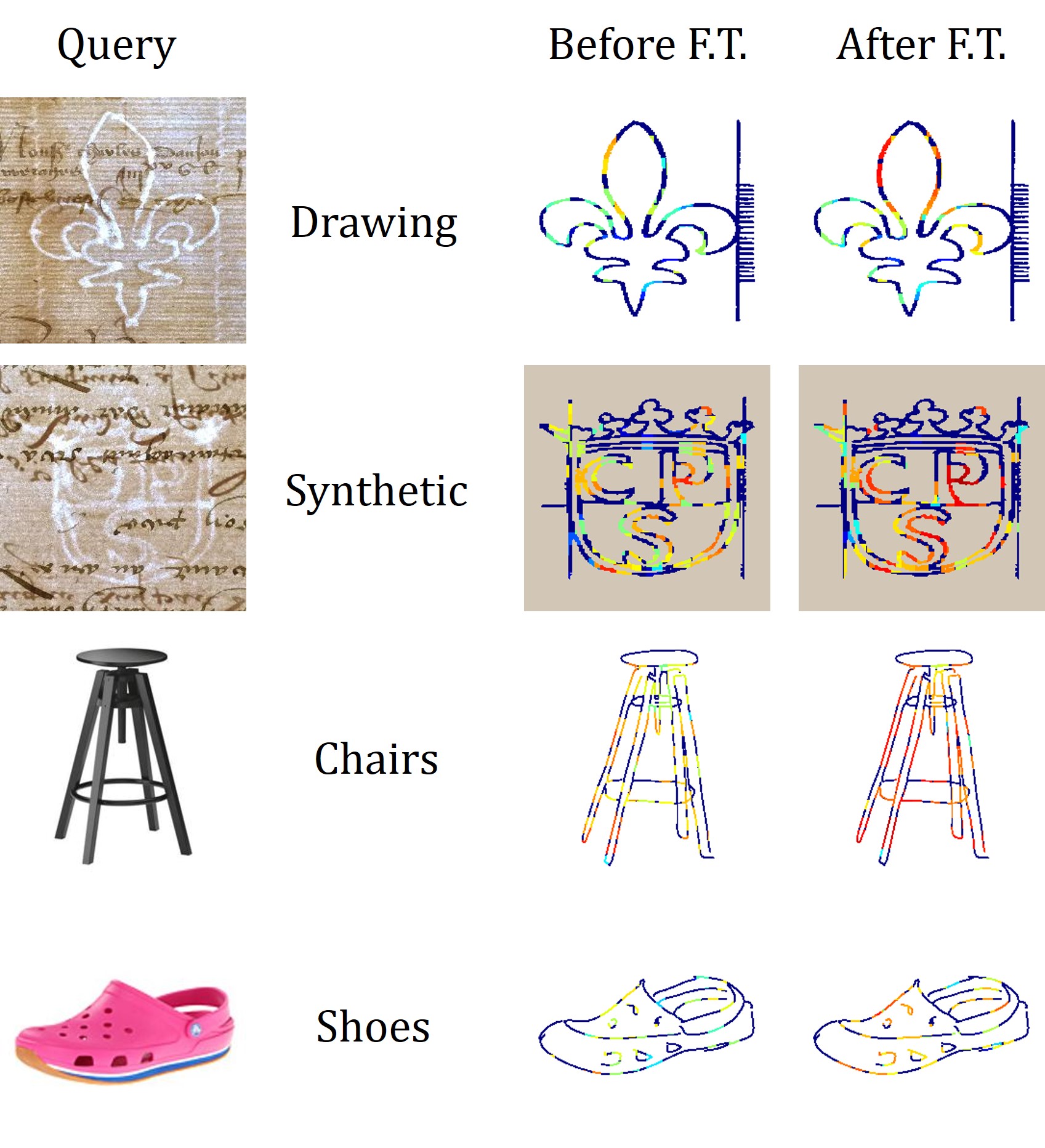}
 	\vspace{-2mm}
 	\caption{Visualization of the effect of fine-tuning. The color is set to red if the local score is 1, blue if it is 0. Notice how more regions contribute after fine-tuning and how the contributions are stronger.}
 	\label{fig:part2Vis}
 	\vspace{-4mm}
\end{figure}

\begin{figure}[t]
	\centering
  	\includegraphics[width=0.18\linewidth]{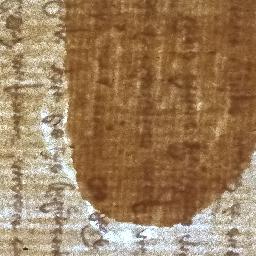}
  	\includegraphics[width=0.18\linewidth]{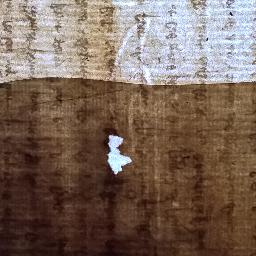}
  	\includegraphics[width=0.18\linewidth]{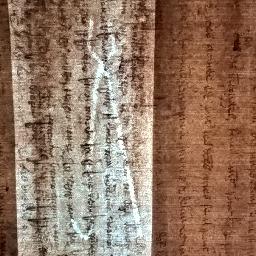}
  	\includegraphics[width=0.18\linewidth]{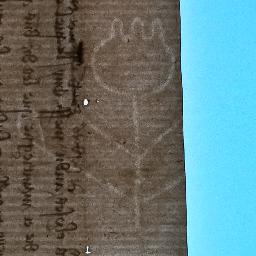}
  	\includegraphics[width=0.18\linewidth]{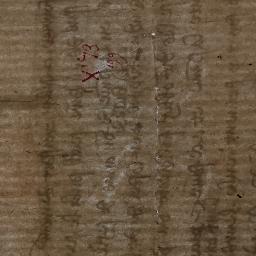}
  	\\[1mm]
  	\includegraphics[width=0.18\linewidth]{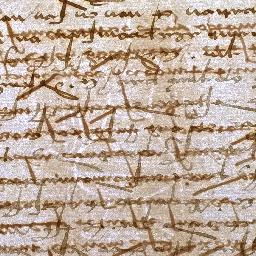}
  	\includegraphics[width=0.18\linewidth]{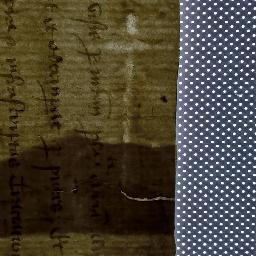}
  	\includegraphics[width=0.18\linewidth]{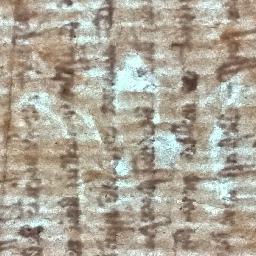}
  	\includegraphics[width=0.18\linewidth]{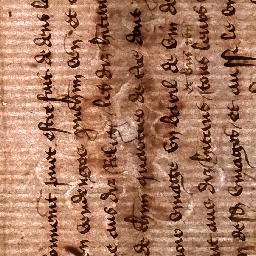}
  	\includegraphics[width=0.18\linewidth]{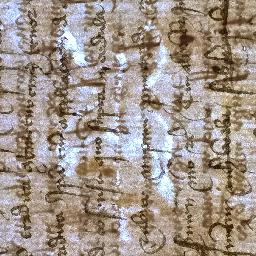}
  	\vspace{-2mm}
 	\caption{Typical failure examples: strong lighting effect, stain or cut on the paper, barely visible watermarks.}
 	
 	\label{fig:failureCase}
 	\vspace{-4mm}
 	
\end{figure}

\paragraph*{Results and failure modes} On the project website, we provide top-5 matches on dataset-B using the 16,753-class curated Briquet catalog as reference. One can first notice the ability of our approach to select the correct class even when very similar classes exist. Failures are often easy to understand, we show typical examples in Figure~\ref{fig:failureCase}. We identified two main types of failures. First, some failures are related to effects in the images that were rare in our dataset A, such as strong lighting effects, stains on the paper, or even tears. We believe this could be remedied using specific data augmentation. Second are cases where the watermark is not easily visible in the photograph, which are expected.


\section{Conclusion}
\IEEEPARstart{W}{e} have identified several challenges for the practical application of automatic watermark recognition. We overcame the difficulty of data collection and, to the best of our knowledge, we present the first publicly available dataset of watermarks providing many photographs of different instances for a large number of classes. We propose a new image similarity and feature fine-tuning strategy, improving over state-of-the-art deep learning approaches for one-shot and cross-domain watermark recognition, and providing strong results for one-shot fine-grained cross-domain 16,753-class watermark classification.

\section*{Acknowledgment}

\IEEEPARstart{T}{his} work was partly supported by ANR project EnHeritANR-17-CE23-0008 PSL "Filigrane pour tous "\footnote{\url{https://filigranes.hypotheses.org/english}} project and gifts from Adobe to \'Ecole des Ponts.

\ifCLASSOPTIONcaptionsoff
  \newpage
\fi

\bibliographystyle{IEEEtran}
\bibliography{bibtex/bib/IEEEabrv,bibtex/bib/IEEEexample}{}

\end{document}